\documentclass[10pt,twocolumn,letterpaper]{article}

 \usepackage[pagenumbers]{cvpr} 

\usepackage{algorithm}
\usepackage{algorithmicx}
\usepackage{algpseudocode}
\usepackage{amsmath}

\definecolor{cvprblue}{rgb}{0.21,0.49,0.74}
\usepackage[pagebackref,breaklinks,colorlinks,allcolors=cvprblue]{hyperref}
\usepackage{multirow}
\usepackage{graphicx}
\usepackage{array}
\usepackage[table]{xcolor}
\usepackage{longtable}

\def\paperID{\*\*\*\*\*\*} 
\def\confName{CVPR}
\def\confYear{2026}

\title{SpikeTrack: A Spike-driven Framework for Efficient Visual Tracking }

\author{
\parbox{\textwidth}{\centering
Qiuyang Zhang\textsuperscript{1}, 
Jiujun Cheng\textsuperscript{1,\textdagger}, 
Qichao Mao\textsuperscript{1,\textdagger}, 
Cong Liu\textsuperscript{2},
Yu Fang\textsuperscript{1},
Yuhong Li\textsuperscript{3},
Mengying Ge\textsuperscript{4},
Shangce Gao\textsuperscript{5}\\[6pt]
\textsuperscript{1}Tongji University \quad
\textsuperscript{2}Nova University of Lisbon \quad
\textsuperscript{3}Stockholm University\\[2pt]
\textsuperscript{4}Shanghai University \quad
\textsuperscript{5}University of Toyama
}
}

\begin{document}

 \maketitle
 \begin{abstract}

Spiking Neural Networks (SNNs)  promise energy-efficient vision, but applying them to RGB visual tracking remains difficult: Existing SNN tracking frameworks either do not fully align with spike-driven computation or do not fully leverage neurons’ spatiotemporal dynamics, leading to a trade-off between efficiency and accuracy.  To address this,  we introduce SpikeTrack, a  spike-driven framework  for  energy-efficient RGB object  tracking.  
SpikeTrack employs a novel asymmetric design that uses asymmetric timestep expansion and unidirectional information flow, harnessing  spatiotemporal dynamics while cutting computation.  To ensure effective unidirectional information transfer between branches, we design a memory-retrieval module inspired by neural inference mechanisms. This module recurrently  queries a compact memory initialized by the template to retrieve target cues and sharpen target perception over time.  Extensive experiments demonstrate that SpikeTrack achieves the state-of-the-art among SNN-based trackers and remains competitive with advanced ANN trackers.  Notably, it surpasses TransT on LaSOT dataset while consuming only 1/26 of its energy. To  our knowledge, SpikeTrack is the first spike-driven framework to make RGB tracking both accurate and energy efficient.  The code and models are available at this \href{https://github.com/faicaiwawa/SpikeTrack}{ URL}.

\end{abstract}

\begingroup
\renewcommand{\thefootnote}{}
\makeatletter
\renewcommand{\@makefntext}[1]{\noindent #1}
\makeatother
\footnotetext{\textsuperscript{\textdagger} Corresponding authors: Jiujun Cheng (chengjj@tongji.edu.cn), Qichao Mao (mao\_qichao@tongji.edu.cn).}
\endgroup

 \section{Introduction}
\label{sec:intro}

Spiking neural networks (SNNs) are a promising energy-efficient computing paradigm that simulates the spatiotemporal dynamics and spiking mechanisms of biological neurons~\cite{snn1}. Their spike-driven computation has two advantages: (i) computation is triggered only when driven by events~\cite{snn2_event}, and (ii) matrix multiplications between spike tensors and  weights can be converted into sparse additions~\cite{snn_mul2add}. This gives SNNs a significant power-saving advantage over ANNs on neuromorphic chips~\cite{snn_powersave1,snn_powersave2}. SNNs have shown strong results on multiple vision tasks~\cite{spikeyolo,spikevideoformer,spike2former}, and their spatiotemporal dynamics make them natural candidates for tracking continuously moving objects.

\begin{figure}
    \centering
    \includegraphics[width=1\linewidth]{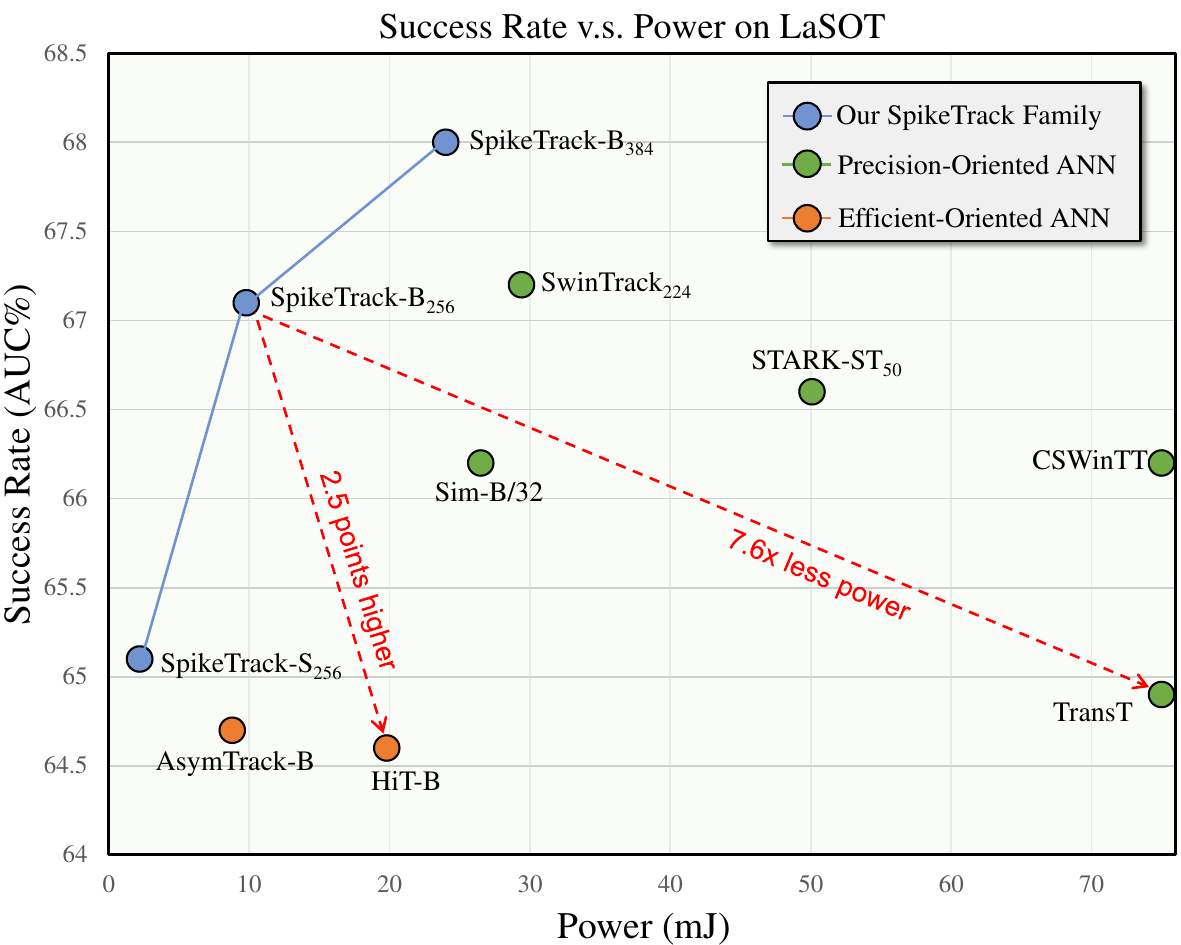}
    \caption{Energy–accuracy trade-off  on LaSOT\cite{lasot}. SpikeTrack achieves lower energy consumption than efficient ANN trackers while matching the accuracy of precision-oriented methods. }
    \vspace{-10pt} 
    \label{fig:bubble}
\end{figure}

\begin{figure}
    \centering
    \includegraphics[width=1\linewidth]{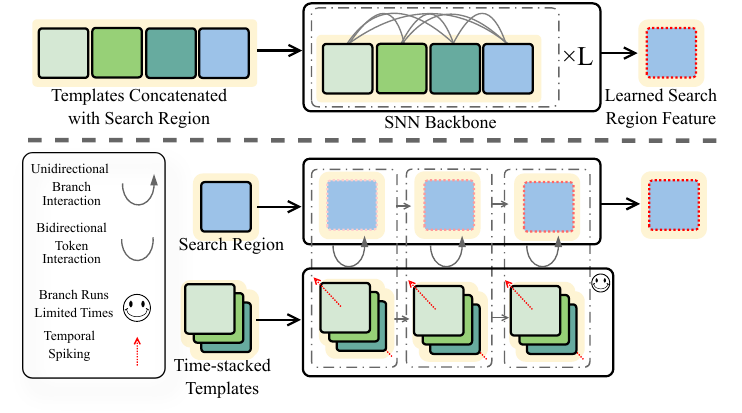}
    \caption{Structure comparison  between one-stream tracking SNN (top)  and our asymmetric tracking SNN (bottom). \textit{L} represents the number of blocks in the backbone. }
\vspace{-10pt} 
    \label{fig:frame_com}
\end{figure}

Current SNN tracking work falls into RGB-based and event-based methods. Within the RGB-based line, SiamSNN~\cite{siamsnn} and Spike-SiamFC++~\cite{spikesiamfc}, adopt the Siamese architecture, achieve tracking via network conversion and end-to-end training, respectively. Although these methods use spiking neurons in form, they decode spike signals into continuous values for computation, preventing fully spike-driven processing and reducing energy efficiency. Event-based methods~\cite{sdtrack,sdtrack2} adapt the dense interaction framework~\cite{ostrack,simtrack} from ANN, also known as the one-stream architecture, as shown in Fig.~\ref{fig:frame_com}. This approach concatenates the search region and  multiple templates along the token length dimension within a single timestep, feeding them into the backbone for joint modeling via spike self-attention. However, this direct imitation underuses the spatiotemporal associative dynamic  of SNNs, and dense,  bidirectional interactions greatly increase computational overhead. This raises a research question: Can we design an SNN that adheres to the spike-driven paradigm while fully leveraging spatiotemporal modeling capabilities for efficient RGB tracking?

To address this problem, we propose SpikeTrack, a spike‑driven SNN for energy‑efficient RGB tracking. SpikeTrack adopts an asymmetric Siamese architecture, with asymmetric timestep inputs and unidirectional information transfer, as shown in Fig.~\ref{fig:frame_com}. Specifically, the template branch expands across multiple timesteps, assigning a template to each step and jointly modeling template representations through neuron's spatiotemporal  dynamics, while the search branch performs efficient single-timestep inference. Information flows only from the template branch to the search branch, allowing the computation-heavy template branch to run only during initialization or template updates, thereby cutting computation.  Additionally, to ensure effective unidirectional information transfer between branches, we design a memory-retrieval module (MRM) inspired by neural inference mechanism~\cite{recurrent}. This module recurrently queries a compact memory initialized from the template features to retrieve target cues and sharpen target perception over time.

Extensive experiments demonstrate that SpikeTrack achieves strong energy efficiency and accuracy with a simple framework, outperforming prior  SNN-based trackers. For instance, SpikeTrack-S$_{256}$ outperforms SpikeSiamFC++ by 8.5\% on the UAV123 dataset. Moreover, as shown in Fig.~\ref{fig:bubble}, SpikeTrack-S$_{256}$ surpasses the efficiency-oriented AsymTrack~\cite{asymtrack} with 2.5× better energy efficiency, while SpikeTrack-B$_{256}$ outperforms the precision-oriented  TransT~\cite{transt} with 7.6× energy savings and 2.2\% higher accuracy.

Our main contributions are summarized as follows:
\begin{itemize}

\item We design an asymmetric SNN that fully utilizes  the  spatiotemporal dynamics of neuron   while significantly reducing computational cost.

\item We propose a brain-inspired memory retrieval module that enables effective unidirectional information transfer. 

\item Building on the above  designs, we propose SpikeTrack, a spike-driven framework for efficient RGB-based tracking, with a family of model variants.  Experiments across multiple benchmarks demonstrate its effectiveness.

\end{itemize}

 \section{Related Work}
\label{sec:related_work}

\textbf{SNNs in Vision Tasks.} Recently, SNN-based approaches have achieved performance comparable to ANNs across various vision tasks, including image classification~\cite{sdtv1,sdtv3}, object detection~\cite{spikeyolo}, semantic segmentation~\cite{spike2former}, and video classification~\cite{spikevideoformer}, as well as higher-level applications such as autonomous driving perception~\cite{zhu2024autonomous} and embodied intelligence~\cite{robot}. By modeling neuronal membrane potential dynamics, SNNs possess powerful spatiotemporal encoding capabilities, making them particularly promising for tracking tasks that require perceiving continuously moving objects.

\noindent\textbf{Visual Tracking Architecture.}Visual tracking aims to predict a target's position and scale across video frames given its initial template. ANN-based trackers follow either two-stream (Siamese) or one-stream designs. Two-stream methods extract template and search features separately, then model their relation via cross-correlation or Transformer interaction. OSTrack~\cite{ostrack} adopts a one-stream design, concatenating template and search patches in a Vision Transformer to jointly extract and relate features with strong results. However, AsymTrack~\cite{asymtrack} shows that such bidirectional interactions are costly on edge devices, and proposes an asymmetric Siamese network with unidirectional template modulation for competitive lightweight tracking. Inspired by this, we design an asymmetric architecture for RGB-based SNN tracking, using asymmetric timestep inputs and memory-retrieval-based unidirectional transfer to achieve efficient tracking with minimal overhead.

\begin{figure*}[htb!]
    \centering
    \includegraphics[width=1\linewidth]{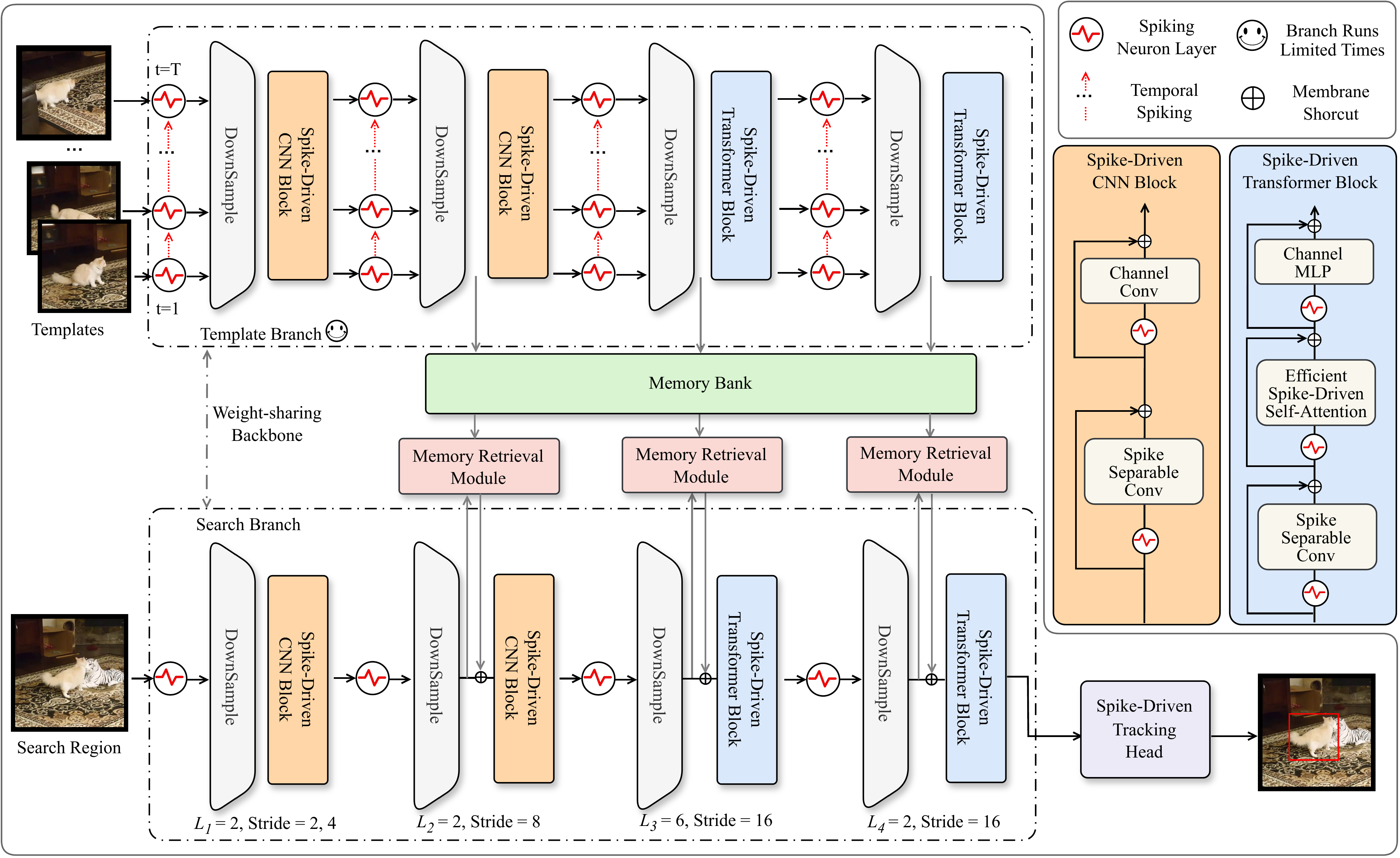}
    \caption{Overview of SpikeTrack. The network consists of three components: a weight-sharing siamese backbone, a memory retrieval module for information transfer, and a prediction head. We use asymmetric timestep inputs and unidirectional information flow. During inference, template branch features are converted and cached as memory. The search branch queries this memory to extract target cues. The template branchs runs once, per initialization or  update. }
    \vspace{-10pt} 
    \label{fig:framework}
\end{figure*}

\noindent\textbf{SNN-based Visual Tracking.}\hspace{0.5em}
Current SNN-based tracking research primarily targets event-camera inputs, where sparse event data and one-stream architectures yield strong results~\cite{sdtrack,sdtrack2}, but the reliance on dedicated hardware limits practical adoption. RGB-based tracking offers a more deployable alternative; however, existing efforts such as SiameseSNN~\cite{siamsnn} and SpikeSiameseFC++~\cite{spikesiamfc} are tied to specific ANN frameworks, suffering from poor scalability, limited performance, and lacking comprehensive evaluation or energy analysis. To address these issues, we propose SpikeTrack, a concise and efficient RGB tracking baseline with extensive benchmark evaluation and detailed theoretical energy analysis.

 \section{SpikeTrack-based Visual Tracking}
\label{sec:method}

\indent In this section, we present the proposed SpiketTrack in detail. We begin with Sec.~\ref{sec:ov}, briefly describing the overall network architecture, followed by Sec.~\ref{sec:snm}, which introduces the basic spiking neuron model used. Subsequently, Sec.~\ref{sec:arch} describe the network components in detail. Finally, Sec.~\ref{sec:tai} introduces the training and inference pipeline.

\subsection{Overview}
\label{sec:ov}

\indent As shown in Fig.~\ref{fig:framework}, SpikeTrack comprises three components: a shared-weight spiking backbone, a memory retrieval module (MRM) for unidirectional branch interaction, and a prediction head. During inference, the template branch performs inference once after template initialization or template update, caching features from different intermediate layers in the memory bank as memory. The search branch then uses MRMs to retrieve the targets cues from  memory and  progressively refine target  perception. Finally, the prediction head consumes the region features to produce the tracking results.

\subsection{Spiking Neuron Model}
\label{sec:snm}

\indent We adopt the Normalized Integer Leaky Integrate-and-Fire (NI-LIF) neuron~\cite{spike2former}. It trains with normalized integer activations based on the classical LIF neuron~\cite{lif}, and converts integer activations into equivalent spikes during inference to preserve spike-driven characteristics. In this work, we design the leaky factor as a trainable variable to allow the network to adaptively model the correlation between timesteps. The neural dynamics equation for NI-LIF is:
\begin{align}
U[t] &= \beta_t H[t-1] + Y[t] \\
S[t] &= \text{Clip}\big(\text{round}\big(U[t]\big),0,D\big)/D\\
H[t] &= U[t] - S[t]\times D\\
\beta_t &= \sigma(\theta_t) 
\end{align}
where $t$ is the timestep, $U[t]$ represents the membrane potential after charging but before firing. Spatial input $Y[t]$ is extracted from the original spike input through a Conv or MLP operation, temporal input $\beta H[t - 1]$ is derived from the decay of the membrane potential at the previous timestep, $\beta_t$ is the leaky factor, $\sigma(\cdot )$ is the sigmoid function, $\theta_t$ is a learnable variable, $S[t]$ is the output spike, $H[t]$ is the membrane potential after firing, $round(\cdot )$ is a round operation, $Clip(x, min, max)$ implies clipping the input $x$ to $[min, max]$, and $D$ is a hyper-parameter to emit the maximum integer value.

\subsection{SpikeTrack Architecture}
\label{sec:arch}

\textbf{Asymmetric Siamese Backbone.} 
We adopt Spike-Driven Transformer v3~\cite{sdtv3} as the backbone. It is a meta-Transformer style SNN~\cite{yu2022metaformer}  composed of convolutional and Transformer-based SNN blocks. Specifically, the input of the backbone are template image $Z \in\mathbb{R}^{T\times3\times H \times W}$, where $T$ depends on the number of templates, and the search image $X \in\mathbb{R}^{3\times H \times W}$.   The input images first pass through a 7 $\times$ 7 convolutional layer for 2$\times$ downsampling, followed by a 3$\times$3 convolutional layer after each stage for 2$\times$ downsampling. Each stage contains \(L_i\) blocks, where the first two stages utilize CNN-based SNN blocks, and the latter two stages incorporate Transformer-based SNN blocks.  
In the template branch, from the second downsampling layer onward, features from each downsampling stage, the intermediate layer of stage3 ($l_3/2$) , and the final layer of stage 4 are initialized in the memory bank.  When the search branch reaches layers aligned with the cached memory, the MRM retrieves target cues from the corresponding memory to enhance target perception. After the final stage, the enriched search feature are fed into the tracking head for prediction.

Next, we introduce the two basic components that make up the backbone:

\noindent\textit{CNN Block.} Each CNN block consists of a spike separable convolution followed by a channel-wise convolution, as detailed below:
\begin{align}
U' &= U + \text{SSConv}(U), \\
U'' &= U' + \text{ChannelConv}(U'), \\
\text{SSConv}(U) &= \text{Conv}_{\text{pw}}(\mathcal{SN} (\text{Conv}_{\text{dw}}(\mathcal{SN} (\nonumber\\
 \quad\; &\quad \text{Conv}_{\text{pw}}(\mathcal{SN} (U)))))), \\
\text{ChannelConv}(U') &=\text{Conv}(\mathcal{SN} (\text{Conv}(\mathcal{SN}(U')))), 
\end{align}
where $\mathcal{SN}(\cdot )$ is the spike neuron layer, $\text{SSConv}(\cdot )$ is the spike separable  convolution, $\text{Conv}(\cdot)$ is the vanilla convolution, $\text{Conv}_{\text{pw}}(\cdot )$ and $\text{Conv}_{\text{dw}}(\cdot )$ are the point-wise  and depth-wise convolutions,  respectively. The BN layers are omitted for brevity.

\noindent\textit{Transformer Block.}  Each Transformer block contains a separable convolution, an efficient spike-driven self-attention module (E-SDSA) and a channel MLP , as detailed below:
\begin{align}
U' &= U + \text{SSConv}(U),\\
U'' &= U' +  \text{E-SDSA}(U'),\\
U''' &= U''+ \text{ChannelMLP}(U''), \\ 
\text{ChannelMLP}(U'') &=\text{Linear}(\mathcal{SN} (\text{Linear}(\mathcal{SN}(U'')))),
\end{align}
where E-SDSA$(\cdot)$ employs binary spiking tensors $K_S$, $Q_S$, $V_S$ $\in$ $\left \{ 0,1 \right \}^{N \times D} $  as the Query, Key and Value, respectively, where $N$ is the token length and $D$ is the channel size. The subscript $S$ represents that the tensor is in spike form. By omitting the softmax function, E-SDSA enables computational order rearrangement to achieve linear complexity with respect to token length. The process is described as follows:
\begin{equation}
Q_S = \mathcal{SN}(\text{Linear}(U)),
\end{equation}
\begin{equation}
K_S = \mathcal{SN}(\text{Linear}(U)),
\end{equation}
\begin{equation}
V_S = \mathcal{SN}(\text{Linear}_{\gamma}(U)),
\end{equation}
\begin{equation}
\begin{split}
U' &= \text{Linear}_{\frac{1}{\gamma} }\mathcal{SN}(\underbrace{(Q_SK_S^{T})V_S}_{\mathcal{O}(N^2D)} * scale) \\
&= \text{Linear}_{\frac{1}{\gamma} }\mathcal{SN}(\underbrace{Q_S(K_S^{T}V_S)}_{\mathcal{O}(ND^2)} * scale)
\end{split}
\label{eq:U_prime}
\end{equation}
where $\gamma$ is the expansion factor of $V_S$, used to enhance the representation of E-SDSA, and the constant $scale$ factor is used for gradient stabilization.

\begin{figure}
    \centering
    \includegraphics[width=1\linewidth]{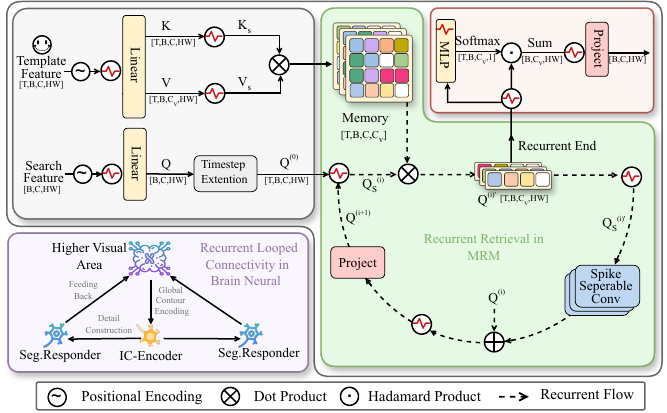}
    \caption{Implementation details of the Memory Retrieval Module. The purple legend (bottom left) illustrates the recurrent,  looped connectivity structure in the brain. For simplicity of illustration, the temporal spiking across timesteps are omitted.}
\vspace{-10pt} 
    \label{fig:mrm}
\end{figure}

\noindent\textbf{Memory Retrieval Module (MRM).} As illustrated in Fig.~\ref{fig:mrm}, the MRM enables effective unidirectional information transfer from the template to the search branch. Its design draws on neuroscientific findings~\cite{recurrent} regarding visual perception under occlusion, where recurrent connectivity in the brain's V1 L2/3 area achieves complete perceptual inference through iterative refinement based on prior expectations—a mechanism naturally aligned with template-based tracking.

For efficiency, features entering the MRM are downsampled via average pooling to the resolution of the final backbone stage and upsampled back at the output. The template feature $F_Z$ is projected to produce Key $K_S$ and Value $V_S$, while the search feature $F_X$ is temporally expanded to produce Query $Q_S^{(0)}$. Leveraging the linear complexity of spike-based attention (Eq.~\eqref{eq:U_prime}), a memory matrix $M = K_S^T V_S$ is pre-computed once during template initialization and reused across frames. From this point, the recurrent loop begins.

The recurrent processing comprises three stages. First, \textit{global contour encoding}: $Q_S^{(i)}$ retrieves from $M$ via a scaled dot-product followed by a spiking neuron:
\begin{equation}
    Q_{S}^{(i)}{'} = \mathcal{SN}(Q_S^{(i)} M \cdot scale).
\end{equation}
Second, \textit{detail construction}: ${Q_S^{(i)}}{'}$ is processed by $T$ dedicated SSConvs along the temporal dimension, with each timestep assigned its own operator to improve sensitivity to temporal variations:
\begin{equation}
    {Q}^{(i)}{''}= \text{SSConv}_t(Q_S^{(i)}{'}[t]), \quad t \in \{1, 2, \ldots, T\}.
\end{equation}
Third, \textit{feedback refinement}: a residual connection with projection simulates feedback to higher-level visual areas:
\begin{equation}
    Q_S^{(i+1)} = \text{Project}(\mathcal{SN}(Q^{(i)} + Q^{(i)}{''})).
\end{equation}
This process repeats for $N$ iterations ($N{=}1$ for all variants). Finally, a spike-driven two-layer MLP generates channel-wise weights $w$ for temporal fusion, and the result is projected back to the original channel dimensions:
\begin{equation}
    F_{out} = \text{Project}\left(\sum_{t=1}^{T} w_t \odot Q_S^{(N)}[t]\right).
\end{equation}

\noindent\textbf{Prediction Head.} We employ a center head to predict the object bounding box, following the design of OSTrack\cite{ostrack} while adopting a spike-driven mechanism. The feature of the search branch are passed through three parallel branches, each composed of several Conv-BN-NILIF layers. The last layer does not contain BN and NI-LIF. These branches predict (1) the target’s center localization  (classification), (2) the local offset induced by resolution reduction, and (3) the normalized bounding-box width and height.

\subsection{Training objective and Inference}
\label{sec:tai}

\textbf{Trainning.} We comine weighted focal loss~\cite{weightedloss}, $\ell_1$ loss and generalized IoU loss~\cite{giou}  as the training objective. The loss function can be formulated as:
\begin{equation}
\mathcal{L }  = {\mathcal{L }_{class}} + {\lambda_G {\mathcal{L }_{IoU}}} + {\lambda_{L_1} {\mathcal{L }_{1}}},
\end{equation}
where ${\mathcal{L }_{class}}$ denotes the weighted focal loss used for classification, ${\mathcal{L }_{IoU}}$ represents the generalized IoU loss, $\ell_1$ is the $\mathcal{L }_{1}$ regression loss, and $\lambda_{G}$ = 2 and $\lambda_{L_1}$ = 5 are the regularization parameters.

\noindent\textbf{Inference.} During the inference process, the template set is regarded as a queue and is updated in a first-in-first-out order while keeping the first initial template fixed. The update strategy follows standard practices~\cite{stark}, using two hyperparameters: a update interval and an update score threshold. The update operation is performed when the update interval arrives and the predicted quality score is higher than the threshold. All models use the same set of hyperparameters.

To reduce training burden and keep the network simple, SpikeTrack omits a separate quality scoring module and instead uses the localization branch score in the prediction head as the confidence score.

 \section{Experiments}
\label{sec:exp}

\subsection{Implementation Details}
\indent  The SpikeTrack models are implemented using Python 3.12 with PyTorch 2.0.0 and trained on 8 NVIDIA 4090 GPUs. 

\noindent\textbf{Model.} We develop six SpikeTrack model variants to balance power and accuracy, varying in backbone size (base/small), input resolution (256/384), and number of timesteps (1/3). We adopt Spike-Driven Transformer (SDT) V3-19M~\cite{sdtv3}  as  backbone for SpikeTrack-Base and SDTV3-5.1M for SpikeTrack-Small. The backbones are initialized with the Imagenet-1K~\cite{deng2009imagenet} pre-trained parameters.

\begin{table*}
  \centering
    \caption{Comparison of performance and efficiency across tracking methods. \textit{Para.} and \textit{Pow.} denote parameters (M) and power ($mJ$), respectively; \textit{T} and \textit{D} represent the timestep and maximum integer value emitted during training. Results marked with $^{*}$ are trained solely on the GOT-10K set. All results are reported in percentage (\%). The top two snn results are highlighted in \textbf{bold} and \underline{underlined}, respectively. }

  \label{tab:tracking_comparison}
  \setlength{\tabcolsep}{3.7pt}
  \renewcommand{\arraystretch}{1.2}
  \begin{tabular}{c|c|>{\centering\arraybackslash}p{0.65cm}%
                >{\centering\arraybackslash}p{0.65cm}%
                >{\centering\arraybackslash}p{0.65cm}|ccc|ccc|ccc|ccc}
    \toprule
    & \multirow{2}{*}{Method} 
    & \multicolumn{3}{c|}{Efficiency}  
    & \multicolumn{3}{c|}{ TrackingNet} 
    & \multicolumn{3}{c|}{ GOT-10k} 
    & \multicolumn{3}{c|}{LaSOT } 
    & \multicolumn{3}{c}{LaSOT$_{ext}$ } \\
    \cline{3-17}
    & & Para. & Pow. & T$\times$D & AUC & P$_{N}$ & P & AO  &SR$_{50}$  & SR$_{75}$  & AUC & P$_{N}$ & P & AUC & P$_{N}$ & P \\
    \cline{1-17}
    \multirow{6}{*}{\rotatebox[origin=c]{90}{SNN}} 
    & \multirow{2}{*}{SpikeTrack-B$_{384}$}  & \multirow{2}{*}{36.8} & 27.3 &3$\times$4  
    &\textbf{82.0}  &\textbf{87.6}  &\textbf{80.7}    & \textbf{73.1}		 & \textbf{84.0} & \textbf{69.9} & 66.7  &76.8  &  \underline{72.9}    & \textbf{47.6} &\textbf{58.5} &\textbf{54.5} \\  
    
    &  &  & 20.0 &   1$\times$4   & \underline{81.8} & \underline{87.1}  & \underline{80.4} & 70.7 & 81.0 & 67.0 & \textbf{67.5}  & \textbf{77.8}  & \textbf{73.3} & \underline{47.3} & \underline{57.8} & \underline{54.1}\\

    \cline{3-17}
    &  \multirow{2}{*}{SpikeTrack-B$_{256}$}  & \multirow{2}{*}{36.8} & 9.8 & 3$\times$4  & 81.1  & 86.9 &  79.1&  \underline{72.2} &\underline{83.9}  &\underline{67.7}  &  \underline{67.1} &\underline{77.7}  & 72.5 &  46.7 & 57.6 & 52.9 \\

    &  &  & 8.1 & 1$\times$4 	&80.1  &85.4  & 77.6 & 69.6 & 80.6 & 66.2 &66.6  & 77.2  &  71.6 & 46.0 & 56.7 &52.0 \\

    \cline{3-17}
    & \multirow{2}{*}{SpikeTrack-S$_{256}$}  & \multirow{2}{*}{11.2} & 3.7 & 3$\times$4   & 78.7 & 84.6 & 75.5 & 67.8 & 79.9 & 61.7 & 64.5  & 76.0 & 69.0   & 43.9 & 54.5 &49.2\\
    
& &  & 2.8 & 1$\times$4   & 77.9  & 83.6 & 74.8 & 67.2 & 78.8 & 60.9  & 65.1  &  76.4   & 69.6 & 43.3 &53.4  &48.2     \\

    \cline{2-17}

    & SiamSNN~\cite{siamsnn}  & - & - & 20   & - & - & - & 31.4 & 32.7 & -  & - & - & - & - &-  &-\\
  
    \cline{1-17}
    \multirow{9}{*}{\rotatebox[origin=c]{90}{ANN}} 
    & AsymTrack-B~\cite{asymtrack}  & 3.36  & 8.3 &-   & 80.0 & 84.5 & 77.4 & 67.7 &76.6  &61.4 & 64.7 &73.0  &67.8 & 44.6 & -  & -    \\
    
    & HiT-B~\cite{hit}  &  42.1 & 19.8  & -  & 80.0 & 84.4 & 77.3 & 64.0 &72.1  &58.1  & 64.6 &73.3  & 68.1 &  44.1&  -&- \\
    
    & CSWinTT~\cite{cswintt}  &  25.1 & 75.4 & -  & 81.9 & 86.7 & 79.5 &  69.4$^{*}$& 78.9$^{*}$ &65.4$^{*}$ &66.2  &75.2  & 70.9  & - & - & -\\
    
    & OSTrack$_{256}$~\cite{ostrack}  &  - &98.9  & -   & 83.1 & 87.8 & 82.0 &  71.0$^{*}$& 80.4$^{*}$ &68.2$^{*}$ &  69.1&78.7  &75.2 & 47.4 & 57.3 &53.3 \\
    
    & SwinTrack$_{224}$~\cite{swintrack}  & 23 &29.4  &-  & 81.1 & - &78.4  & 71.3$^{*}$ & 81.9$^{*}$ & 64.5$^{*}$  &  67.2& - & 70.8 & 47.6 & - & 53.9\\
    
    & Sim-B/32~\cite{simtrack}  &  -& 26.5 &-   & 79.1 & - &83.9  & - & - & -  & 66.2 & 76.1 & - & - & - & -  \\
    
    & STARK-ST$_{50}$~\cite{stark}  & 23.5 &50.1  &-   &81.3  & 86.1 & - & 68.0$^{*}$& 77.7$^{*}$ &62.3$^{*}$ & 66.6 & - & - & - &-  & -\\
    
    & TransT~\cite{transt}  &  17.9&  75.2&  - & 81.4 & 86.7 & 80.3 &72.3  & 82.4 & 68.2 & 64.9 & 73.8 & 69.0 & - & - & -\\
    
    & TrSiam~\cite{trsiam}  &  -& - & -  & 78.1 &82.9  &72.7  &  67.3$^{*}$& 78.7$^{*}$ &58.6$^{*}$ & 62.4 & - & 60.6  & - & - & - \\

    \bottomrule
  \end{tabular}
\end{table*}

\noindent\textbf{Training.} We train on standard SOT datasets: COCO~\cite{coco}, LaSOT~\cite{lasot},  TrackingNet~\cite{trackingnet} and GOT-10k~\cite{got} (excluding 1k sequences from the train split to align with the training data of other trackers). The total batch size is 128. Template and search images are generated by expanding target bounding boxes by a factor of 4.  The AdamW~\cite{adamw} optimizer is used for training. All models use the same training strategy

For T=1 models, we train for 320 epochs using 60k image pairs per epoch. The learning rates are set to 4e-5 for the backbone and 4e-4 for the head and MRMs, with a weight decay of 1e-4. The learning rate is reduced by a factor of 10 after 240 epochs.

For T$>$1 models, the training data consist of image groups containing one search region and T templates. Starting from the pretrained T=1 SpikeTrack weights, we train for 60 epochs with learning rates of 4e-4 for the MRM and learnable decay factor, and 4e-5 for other modules. The learning rate is decreased by 10× after 30 epochs.

\noindent\textbf{Inference.} For simplicity, all models use the same set of hyperparameters. The online template update interval is set to 25, with an update confidence threshold of 0.7 by default. A Hanning window penalty is applied to incorporate positional prior information in tracking, following standard practices~\cite{transt}.

\noindent\textbf{Energy evaluation.} We compare SpikeTrack with SNN and ANN tracking methods, following the energy consumption evaluation criteria used in previous work~\cite{sdtv1,sdtv3,spikeyolo}.  
 The ANN energy cost  is calculated as:  
\begin{equation}
 E_{ANN} = \text{FLOPs} \times E_{MAC},
\end{equation}
 while the SNN energy cost is defined as:
\begin{equation}
 E_{SNN} = \text{FLOPs} \times E_{AC} \times SFR \times T \times D,
\end{equation}
 where $ SFR $ denotes the average spike firing rate, $ T $ is the number of timesteps, and $ D $ is the upper limit of integer activation during training. 
All values are based on 32‑bit floating‑point implementations in 45nm technology~\cite{energyeval}, where $ E_{MAC}$ = 4.6 $pJ$  and $E_{AC}$ = 0.9 $pJ$. More evaluation details are presented in the Supplementary Material.

For SpikeTrack energy analysis, we define the spike firing rate as the average spike rate measured on LaSOT and GOT-10K.  The template-branch energy  is estimated by dividing its total energy  by the update interval.

\subsection{Tracker Comparisons}
We compare our SpikeTrack with  SNN trackers  and ANN trackers on seven widely used tracking benchmarks.

\noindent\textbf{GOT-10K~\cite{got}.}
GOT-10k test set contains 180 videos covering a wide range of common challenges in tracking.  As reported in Tab.~\ref{tab:tracking_comparison}, SpikeTrack-S$_{256}$-T3 achieves a comparable AO score to the state-of-the-art efficient ANN AsymTrack-B while consuming only half the energy. In addition, SpikeTrack-B$_{256}$-T1 delivers a 38.2\% AO improvement over the existing SNN tracker SiamSNN.

\noindent\textbf{LaSOT~\cite{lasot}.}
  LaSOT  is a large-scale long-term tracking benchmark. The test set contains 280 videos with an average length of 2448 frames. The results on LaSOT are presented in Tab.~\ref{tab:tracking_comparison}. SpikeTrack-B$_{256}$-T3 surpasses TransT by 2.2\% in AUC, yet requires less than one-seventh of its energy consumption.  
When comparing different SpikeTrack variants, the performance of S256 and B384 models does not improve as $T$ increases. We attribute this to the higher demand for template precision in long-term tracking that our simple scoring mechanism introduces certain low-quality templates during updates, which in turn undermines predictive accuracy.  

\noindent\textbf{LaSOT$_{ext}$~\cite{lasotext}}.
LaSOT$_{ext}$  is a recently released dataset with 150 video sequences and 15 object classes. Across this dataset, as shown in Tab.~\ref{tab:tracking_comparison}, SpikeTrack variants follow the expected pattern: both higher $T$ values and increased input resolution lead to gradual performance gains.  Notably, SpikeTrack-B$_{256}$-T1 achieves a 1.4\% higher AUC than AsymTrack-B while consuming less energy.   

\noindent\textbf{TrackingNet~\cite{trackingnet}.}
TrackingNet is a large-scale dataset containing 511 videos, which covers diverse object categories and scenes. As reported in Tab.~\ref{tab:tracking_comparison},   when matched in AUC score with SwinTrack$_{224}$, SpikeTrack-B$_{256}$-T3 operates at only one-third of the energy cost.  SpikeTrack-B$_{384}$-T3 reaches an equivalent AUC with just 35\% of the energy consumption of CSWinTT.

\noindent\textbf{TNL2K~\cite{tnl2k}.}
TNL2K is a recently released large-scale dataset with 700 challenging video sequences.
As shown in Tab.~\ref{tab:samlldata}, against the strong one-stream ANN baseline OSTrack$_{256}$, SpikeTrack-B$_{384}$-T3 delivers 0.5\% higher AUC while using less than one-third of its energy.  
Similarly, SpikeTrack-S$_{256}$-T1 achieves 0.5\% better AUC than TransT with a mere 3\% of the energy requirement.

\noindent\textbf{UAV123~\cite{uav123} and OTB100~\cite{otb}.}
Both of these are  small-scale benchmarks including 123 and 100 videos respectively.
The results on these two datasets are  presented in Tab.~\ref{tab:samlldata}.  On the OTB dataset, SpikeTrack-S$_{256}$-T3 outperforms existing SNN-based methods SpikeSiamFC++ and SiamSNN by 5\% and 20.1\% AUC, respectively.  For UAV123, SpikeTrack-B$_{256}$-T3 delivers 10\% higher AUC compared to the best previousSNN results.  

\begin{table}[tb]

\setlength{\tabcolsep}{3.5pt}
\renewcommand{\arraystretch}{1.1}
    \centering
    \caption{Performance comparison of different tracking methods. All evaluation metrics are based on Success Rate (AUC) and  reported in percentage (\%). The top two snn results are highlighted in \textbf{bold} and \underline{underlined}, respectively. }
    \label{tab:samlldata}
    \begin{tabular}{cccccc}
        \toprule
        Method & T$\times$D& Pow. &TNL2K  & OTB  & UAV  \\
        \midrule
        \multirow{2}{*}{SpikeTrack-B$_{384}$} 
         & 3$\times$4 & 27.3 &\textbf{54.8} & \underline{68.9}  & \textbf{68.3}  \\
         & 1$\times$4 & 20.0 &\underline{54.2} & 68.3  &  67.2 \\
         \cline{2-6}
        \multirow{2}{*}{SpikeTrack-B$_{256}$}

         & 3$\times$4 & 9.8 &54.0& \textbf{69.6} &  \underline{67.8}\\
         & 1$\times$4 & 8.1 &53.3 & 68.0 & 67.5  \\
         \cline{2-6}
        \multirow{2}{*}{SpikeTrack-S$_{256}$} 

         & 3$\times$4 &3.7 &52.0  &69.4  &66.2  \\
         & 1$\times$4 &2.8 &51.2 & 67.1 &  66.3\\
        \midrule
        SpikeSiamFC++~\cite{spikesiamfc}&  1& - & - & 64.4 & 57.8  \\
        SiamSNN~\cite{siamsnn}&  20& - & - & 49.3 & -  \\
        \midrule
        AsymTrack-B~\cite{asymtrack} & - & 8.3&- & -  &66.5  \\
        HiT-B~\cite{hit} & - & 19.8 & -  & -  &65.6 \\
        OSTrack$_{256}$~\cite{ostrack} & -&  98.9 &54.3 &  - & 68.3  \\
        SwinTrack$_{224}$~\cite{swintrack} &-& 29.4  &53.0 &  - & -  \\
        Sim-B/32~\cite{simtrack} &-& 26.5 &51.1 & -  & -  \\
        STARK-ST$_{50}$~\cite{stark} &- & 50.1 &54.3  & -  &  68.2 \\
        TransT~\cite{transt} & -& 75.2  &50.7 & 69.4  &  68.1 \\
        TrSiam~\cite{trsiam} &- &-  & - &70.8  &  67.4 \\

        \bottomrule
    \end{tabular}
\end{table}

\begin{table}

\setlength{\tabcolsep}{4pt}
\renewcommand{\arraystretch}{1.2}
    \centering
   
    \caption{Ablation Study on LaSOT and GOT-10k. $\bigtriangleup$  denotes the performance change (averaged over benchmarks) compared with the baseline. $^{*}$ indicates the non-spike version. }
    \label{tab:ana}
    \begin{tabular}{c|ccccc}
        \toprule
        \# & Method &Pow.  & GOT-10K   &  LaSOT & $\bigtriangleup$\\
        \midrule
       1&   Baseline&  8.7 &  71.3  & 66.8  & -  \\
 
        2&  Fine-tuning& 8.7 &  71.6 &  66.9 & +0.2  \\
        
        3&  One-stream  & 22.8 & 70.8 & 65.4 & -0.8 \\
        
        4a & Vanilla Cross-attn &  7.6  &  70.9  &  65.0  & -1.1  \\

        4b &  Modulation & 6.8   & 58.3   & 49.9   &  -14.9 \\
   
        4c &  Modulation$^{*}$ &  10.2  &  64.0  & 58.1   &  -8.0 \\
          
        5 & Mean Fusion  &  8.5 & 71.0 &  66.2&  -0.4  \\
        
        6 & Fixed Decay & 8.9 & 68.9 & 66.0 &  -1.9  \\
        
        7 & Smaller Template &  8.3  & 69.6  &  65.7 & -1.4 \\

        \bottomrule
    \end{tabular}
\end{table}

\subsection{Ablation and Analysis.}

 As shown in Tab.~\ref{tab:ana}, we ablate training methods,  architecture design, and hyperparameter settings. For a fair comparison with the one-stream architecture, the baseline (\#1) is a SpikeTrack-B$_{256}$-T2 trained from scratch, rather than (\#2), which  fine-tunes SpikeTrack-B$_{256}$-T1.  Fine-tuning (\#2) requires fewer epochs and performs better than training from scratch (\#1).

\noindent\textbf{Asymmetric $v.s.$ One-stream.}
Following the structure in \cite{sdtrack, sdtrack2} and keeping the same training settings, we compare the one-stream architecture with our asymmetric architecture, as shown in  Tab.~\ref{tab:ana} (\#3). Our method achieves better results with lower energy consumption. This shows that modeling templates with spatiotemporal neuron dynamics, combined with a memory retrieval module, outperforms  jointly modeling all templates and search regions using with a backbone . 

\noindent\textbf{Effectiveness of the MRM.}
We replace the MRM with vanilla spike cross‑attention, enabling search region features to learn from concatenated template features. As shown in Tab.~\ref{tab:ana} (\#4a), this modification eliminates spatiotemporal processing and loop operations, reducing energy consumption but causing a noticeable drop in accuracy compared to the baseline. Furthermore, we replace the MRM with AsymTrack\cite{asymtrack}'s template modulation module, implementing both spike and non-spike versions. (\#4b) notes that the method is very lightweight after spiking conversion, but suffers severe performance degradation. The hybrid structure of (\#4c) improves performance but remains suboptimal. This indicates that using templates as convolutional kernels for signal modulation is unsuitable for the coarse-grained representation of spiking networks.

\noindent\textbf{Effectiveness of the Fusion Module.}
Table~\ref{tab:ana} (\#5) compares the commonly used time-step averaging fusion method in SNN structures \cite{spikeyolo,sdtv3,sdtv1} with our proposed channel-wise weighted fusion method. The latter performs better.

\noindent\textbf{Learnable Decay  $v.s.$ Fixed Decay.} As shown in Tab.~\ref{tab:ana} (\#6), we compare the fixed membrane potential leaky factor used in previous SNN works~\cite{spikeyolo,sdtv1,sdtv3,spike2former} with our learnable one. The learnable factor enables more flexible and controllable interactions across timesteps.

\noindent\textbf{Template Expand Factor.}
Unlike previous methods, SpikeTrack adopts a template setting with the same size and expansion factor as the search region. Our experiments show that larger template expansion factors and higher template resolutions significantly improve accuracy, as presented in Tab.~\ref{tab:ana} (\#7). We hypothesize that binary tensors lack fine-grained target detail, incorporating background information for contrastive representation allows for more global context and improves target encoding.

\begin{figure}

    \includegraphics[width=1\linewidth]{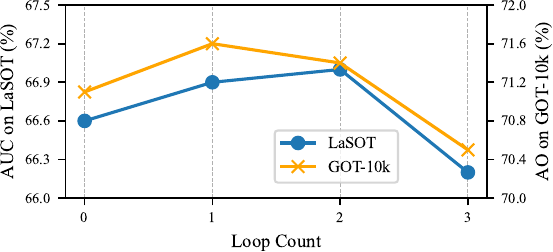}
    \caption{Influence of the number of retrieval loop in MRM.}
    \captionsetup{skip=-1pt}
\vspace{-10pt} 
    \label{fig:loop}
\end{figure}

 \noindent\textbf{Loop Count in MRM.}
Figure~\ref{fig:loop} shows the results obtained with different retrieval loops in MRM. When the number of loops exceeds 1, we add a channel‑wise learnable layerscale on   residual,  to ensure training stability.  One or two loops work best, while more can reduce performance due to accumulated errors and overly narrow focus.

\begin{figure}

    \includegraphics[width=1\linewidth]{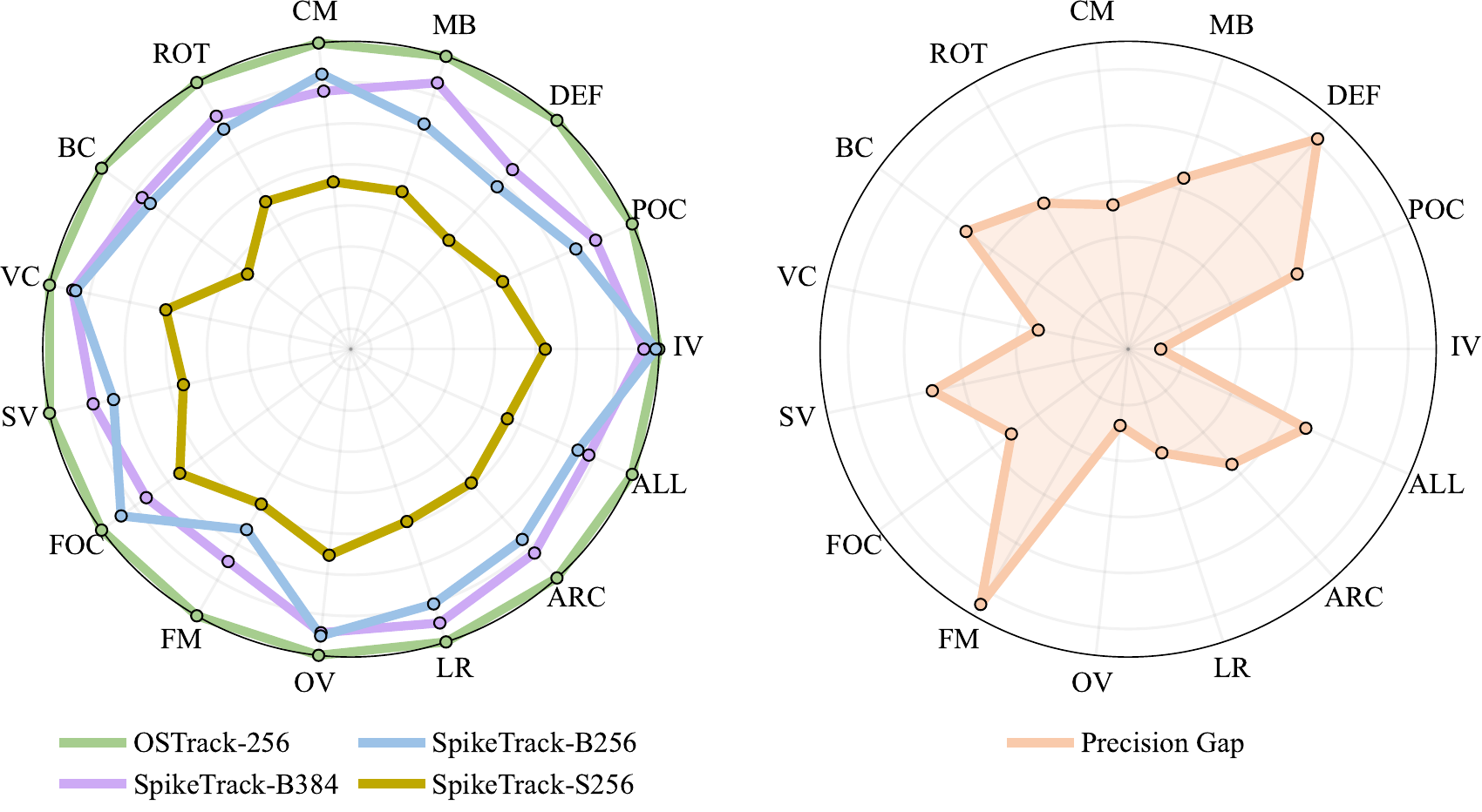}
    \caption{Gap analysis between SpikeTrack and a precision-oriented ANN across LaSOT attributes.}
\vspace{-10pt} 
    \label{fig:lidar}
\end{figure}

\noindent\textbf{Gap Analysis with Precision-Oriented Tracker.} We compare SpikeTrack with the precision-oriented tracker OSTrack~\cite{ostrack} across 14 attributes of the LaSOT dataset, as shown in the left panel of Fig.~\ref{fig:lidar}.  A noticeable performance gap still exists between them. The right panel illustrates the average AUC gap between SpikeTrack-B variants and OSTrack-256. The largest gaps occur in the Deformation and Fast Motion scenarios, which pose greater challenges for deep semantic understanding and re-detection capabilities.  We hope future SNN-based tracker designs can narrow the gap with ANN methods based on these insights. The full names of all attributes are provided in the Supplementary Material.

\begin{figure}

    \includegraphics[width=1\linewidth]{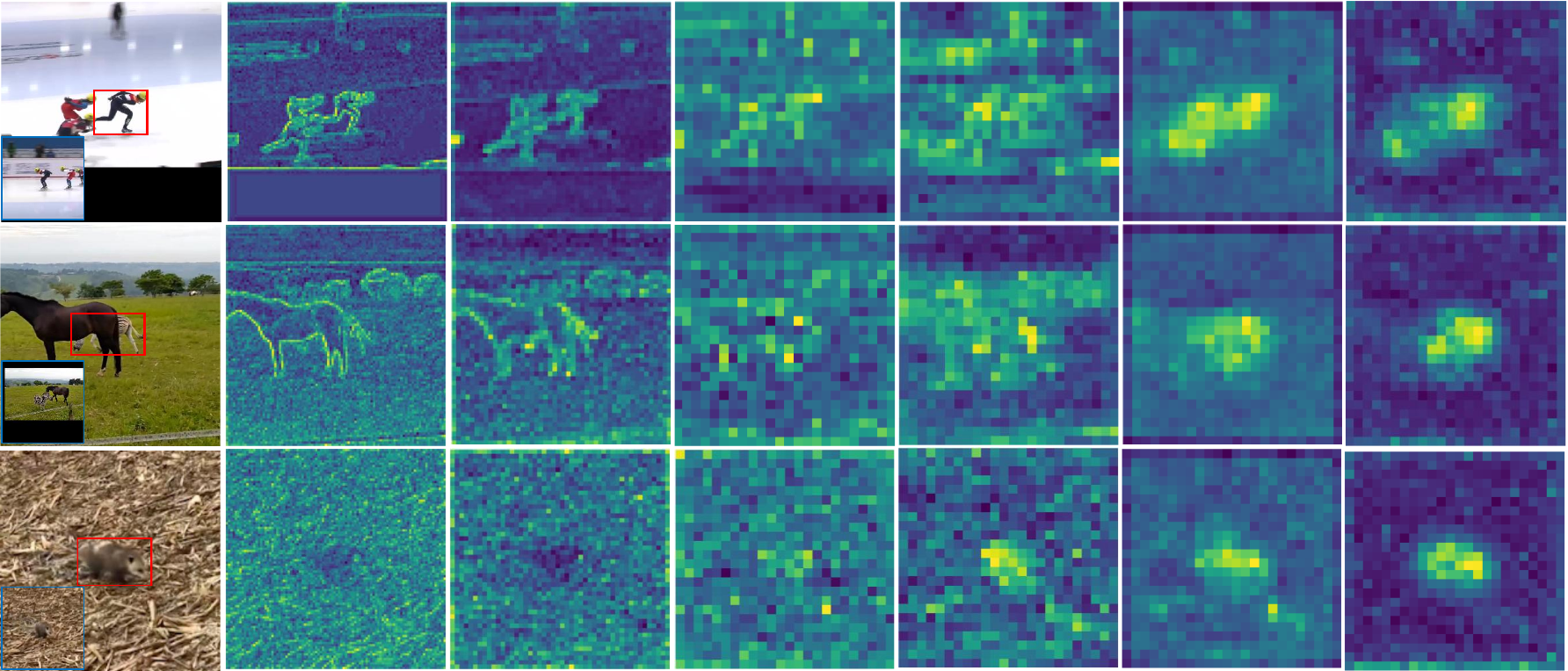}
    \caption{Visualization of the spike tensor produced by MRM. Three cases are shown: similar objects, occlusion, and background interference. }
\vspace{-10pt} 
    \label{fig:visual}
\end{figure}

\noindent\textbf{Visual Analysis.} As shown in Fig.~\ref{fig:visual}, we visualize the spiking outputs of each MRM layer under three challenging scenarios. It can be observed that MRM follows a global-to-instance perception process, building an understanding of the search region based on cues provided by memory. The method performs well under occlusion and background distraction, but in the similar-object interference scenario, although it ultimately locates the correct target, it is still affected by similar objects. We attribute this to the difficulty of representing fine-grained semantic information using spike-based encoding.

 \section{Conclusion}
\label{sec:conclusion}

This work presents SpikeTrack,  a family of spike-driven visual tracking models. With an asymmetric architecture and memory-retrieval-based unidirectional information transfer, SpikeTrack delivers energy-efficient and accurate RGB tracking. Extensive experiments demonstrate that SpikeTrack not only sets a new state-of-the-art among SNN-based trackers, but also shows competitive performance compared to recent ANN-based trackers, while significantly reducing energy consumption.  We hope this work will advance research on SNN for RGB tracking and help narrow the gap with ANN-based trackers.

\noindent\textbf{Limitations.}  A limitation of SpikeTrack lies in its difficulty handling scenes with similar objects. This is because the network does not have explicit modules for distinguishing similar objects, and spike information alone is insufficient to convey the fine-grained representations needed for such distinctions. In future work, we plan to build on this work to explore how to transmit fine-grained representations through spike-based mechanisms to tackle these challenging scenarios.

\section*{Acknowledgments} This work was supported in part by the NSFC under grant 62272344, in part by projects UID/04152/2025 and UID/PRR/04152/2025 from FCT (Fundação para a Ciência e a Tecnologia) through the Centro de Investigação em Gestão de Informação (MagIC)/NOVA IMS.


{
    \small
    \bibliographystyle{ieeenat_fullname}
    \bibliography{main}
}
\clearpage
\setcounter{page}{1}
\maketitlesupplementary

In the supplementary materials, Sec.~\ref{sec:Visualization Results} presents a visual comparison of the tracking results. Sec.~\ref{sec:Attribute} presents SpikeTrack’s performance on each attribute in the LaSOT Benchmark’s attribute challenge. Finally, Sec.~\ref{sec:Power} explains the method for estimating energy consumption and reports the spiking firing rate (SFR) for each layer of SpikeTrack‑B256‑T3.

\section{Visualization}
 \label{sec:Visualization Results}

 \subsection{Visualization Retrieval Results} 
\begin{figure}[h]
    \centering
    \includegraphics[width=0.8\linewidth]{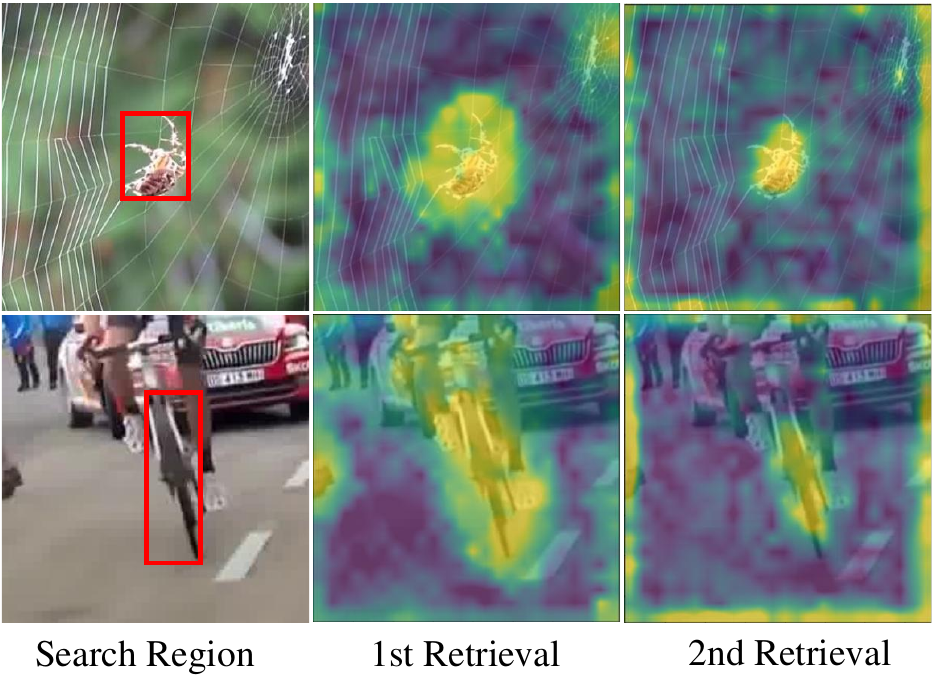}
    \caption{Visualization of the spike tensor obtained after two retrievals in MRM, based on spike firing rate.}
\vspace{-10pt} 
    \label{fig:rb_vis_2}
\end{figure}

To further demonstrate the effectiveness of recurrent retrieval, we visualize the retrieval results at each iteration based on the per-pixel channel spike firing rate. Fig.~\ref{fig:rb_vis_2} demonstrate a progressive attention refinement process, with increasingly concentrated spike firing locations  and reduced noise, validating the recurrent design’s effectiveness.  This result also supports our hypothesis regarding the experimental observations in Fig.~\ref{fig:loop}: excessive recurrent iterations narrow the attention excessively, causing useful information to be overlooked.

\subsection{Visualization Tracking Results} 

\begin{figure*}
    \centering
    \includegraphics[width=1\linewidth]{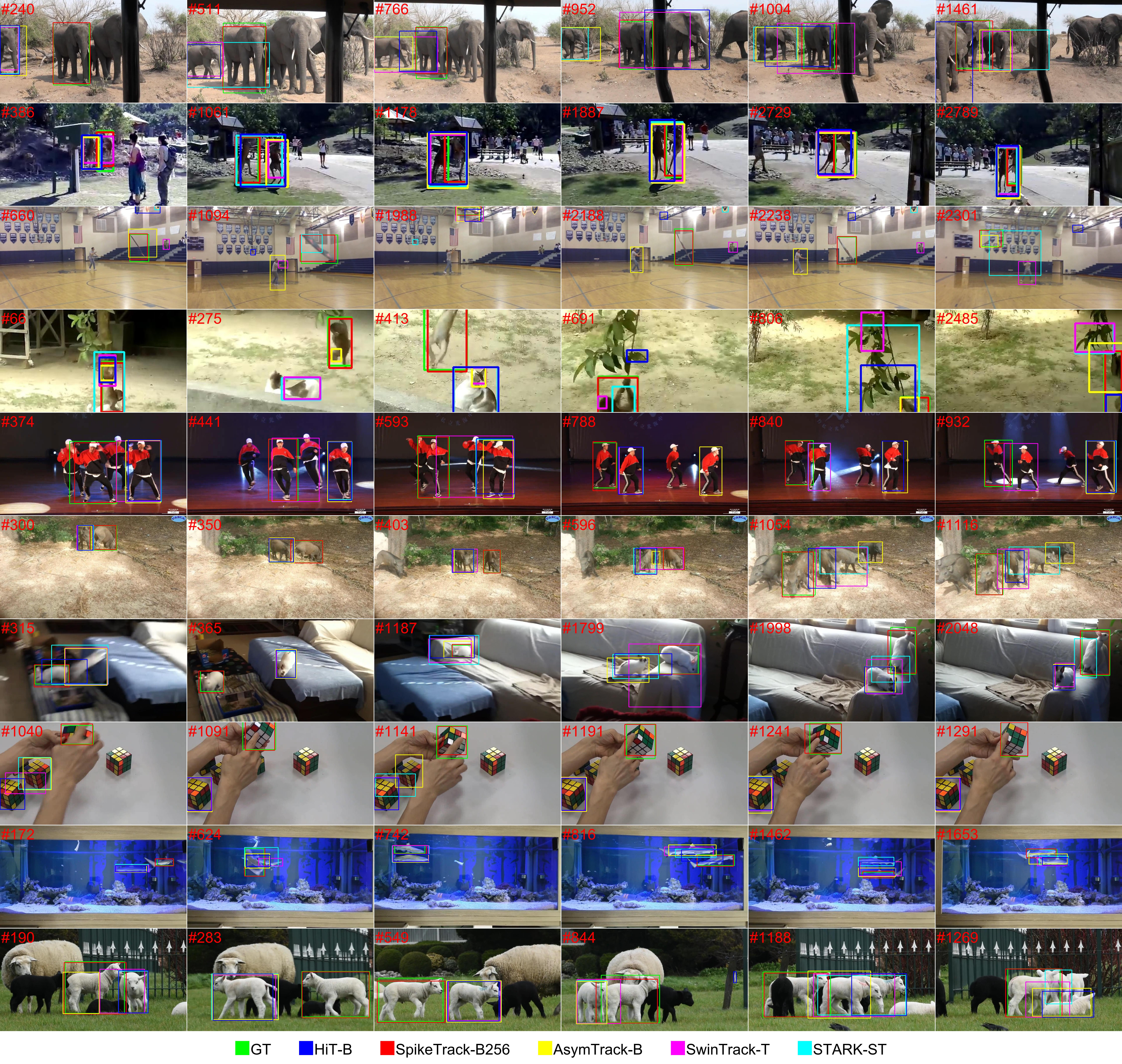}
    \caption{Visualization comparison of SpikeTrack and other ANN-based Trackers. }
\vspace{-10pt} 
    \label{fig:vis}
\end{figure*}

As shown in Fig.~\ref{fig:vis}, we present the tracking results of SpikeTrack, efficiency-oriented ANNs, and precision-oriented ANNs. The video sequences include challenging scenarios such as deformation, blur, and similar objects, demonstrating SpikeTrack’s ability to maintain accurate tracking over extended time spans.

\begin{figure*}
    \centering
    \includegraphics[width=1\linewidth]{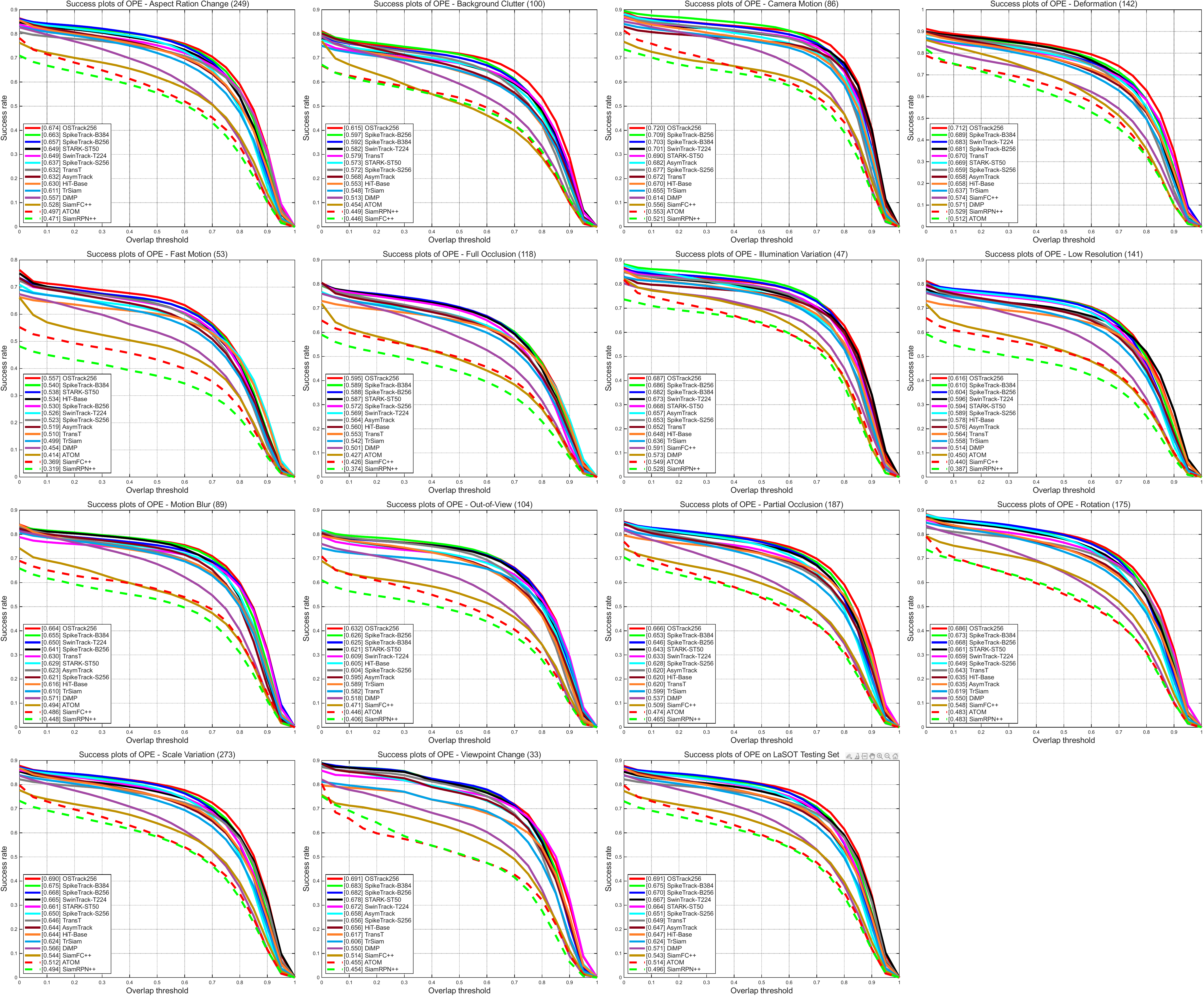}
    \caption{Performance of SpikeTrack and other ANN-based Trackers on the LaSOT attribute challenges. }
\vspace{-10pt} 
    \label{fig:merged_noloss}
\end{figure*}

\section{More Exploring Experiments}  
\subsection{Updating Hyper-parameters}

\begin{table}[htbp]
    \centering
    \setlength{\abovecaptionskip}{-2pt}  
    \setlength{\belowcaptionskip}{-1pt}  
    \caption{Template update parameter sensitivity analysis of SpikeTrack-B$_{256}$-T3. Rows indicate update confidence and columns indicate update interval.  Underlines indicate default unified settings of the SpikeTrack Family for short- and long-term datasets. }
    \label{tab:update}
    \small
    \setlength{\tabcolsep}{3pt}  
    \renewcommand{\arraystretch}{0.7}  
    
    \begin{tabular}{c|cccc}
        \toprule
        GOT&\underline{25}& 35 &45  &55 \\
        \midrule     
        0.65 &72.2 & 72.5 &  72.2& 72.1\\
        \underline{0.7} & 72.2 & 72.2 & 72.0&71.9\\
        0.75& 71.8 &  72.4& 71.6&71.8\\
        0.8&  71.7&  72.3&71.3 &71.7\\
        \bottomrule    
    \end{tabular}
    \hspace{6pt}  
    \begin{tabular}{c|ccc}
        \toprule
        LaSOT&\underline{40}& 60 &80   \\
        \midrule     
        0.65 &  66.8  &  66.8  & 67.1 \\
        0.7 & 67.0 &   67.2 & 67.1\\
        0.75& 66.7 & 67.6 & 66.7\\
        \underline{0.8}& 67.1 & 67.5 &66.8 \\
        \bottomrule    
    \end{tabular}
    \vspace{-10pt}  
\end{table}
We conduct a sensitivity analysis about updating hyper-parameters.   As shown in Tab.~\ref{tab:update}, within reasonable ranges, lower update intervals and thresholds favor short-term tracking, whereas long-term tracking exhibits the opposite trend, suggesting a stronger dependence on stable templates. Note that our default setting isn’t optimal since we use unified settings for  all datasets (except LaSOT) to show true generalization.

\section{Performance of Attribute Challenges on LaSOT}  
\label{sec:Attribute}
As shown in Fig.~\ref{fig:merged_noloss}, we provide a detailed comparison of the scores achieved by SpikeTrack and other ANN tracking methods for each challenge attribute in LaSOT. These challenge attributes include: Illumination Variation (IV), Partial Occlusion (POC), Deformation (DEF), Motion Blur (MB), Camera Motion (CM), Rotation (ROT), Background Clutter (BC), Viewpoint Change (VC), Scale Variation (SV), Full Occlusion (FOC), Fast Motion (FM), Out-of-View (OV), Low Resolution (LR), Aspect Ratio Change (ARC).

\section{Energy Consumption Estimation}  
\label{sec:Power}

\subsection{Spike-driven Operators in SNNs}
Spike‑driven operators are the foundation of low‑power neuromorphic computing, especially for SNNs. In spike‑driven convolution and MLP computations, matrix multiplication between weight matrices and input spike matrices can be implemented on neuromorphic chips as address‑based additions~\cite{snn_mul2add}. In spike‑driven attention mechanisms, the $Q_S$, $K_S$, and $V_S$ operations involve two matrix multiplications. Similar to Conv and MLP, selecting one side as the spike matrix and the other as the weight matrix transforms the computation into sparse additions. Table~\ref{tab:energy} shows a comparison of the energy consumption of ANN and Spike-driven SNN operators.

\subsection{Energy Consumption of SpikeTrack}
In this work, we use the same refined energy consumption evaluation method as recent SNN studies~\cite{sdtrack,sdtrack2,sdtv1,sdtv3,spike2former,spikevideoformer}. We first measure the spiking firing rate (the proportion of non-zero elements in the spike matrix, $SFR$) of each layer, then calculate each layer’s energy consumption as its FLOPs multiplied by \(E_{AC}\) and the corresponding $SFR$, and finally sum the energy consumption across all layers.  In 45nm technology~\cite{energyeval},  the energy of a $MAC$ and an $AC$ are $E_{MAC}$ = 4.6$pJ$ and $E_{AC}$ = 0.9$pJ$ , respectively.

The $SFR$ of each layer is obtained by averaging the firing rates of the  model over the large-scale benchmarks LaSOT and GOT-10K. In order to give readers an intuitive feeling about the spiking firing rate, we give the detailed spiking firing rates of SpikeTrack-B256-T3 in Tab.~\ref{tab:spikingtate_search} (search branch) and  Tab.~\ref{tab:spikingtate_template} (template branch).    Overall, the firing rates exhibit a decreasing trend from shallow to deep layers. The TemporalFusion-WeightMLP1 layer in the MRM module shows a relatively high firing rate, indicating that while the spatiotemporal membrane potential processing and recurrent retrieval introduced by the MRM module improve performance, they also increase energy consumption to some extent.

Notably, the NI-LIF neurons~\cite{spike2former} are trained with normalized integer activation values, with the max integer capped at 4. During inference, these integers are mapped to equivalent spikes, which can naturally result in certain layers exhibiting an $SFR$ greater than 1.  

More SpikeTrack variations of spiking firing rate and energy consumption calculation tables can be found in the Excel document in the supplementary material ZIP.

\begin{table*}
    \centering

    \caption{Energy evaluation. $FL_{Conv}$ and $FL_{MLP} $ represent the FLOPs of the Conv and MLP models in the ANNs, repectively. $R$ denote the spiking firing rate (the proportion of non-zero elements in the spike matrix) of the layer corresponding to the operator.  $T$ is timestep. $E_{MAC}$ and $E_{AC}$ are the energy consumption of performing one MAC and one AC operation, respectively. In the spike-driven attention, the scale operation is avoided by incorporating the scale factor into the neuron’s leakage factor. The first downsampling layer uses $MAC$ convolution, consistent with other works~\cite{sdtv1,sdtv3,spikevideoformer,sdtrack,sdtrack2}.}
    \label{tab:energy}
    \begin{tabular}{cccc}
        \toprule
         &  & ANN-based Tracker  & Spike-driven Tracker    \\
         \midrule
        \multirow{2}{*}{Conv}   & First Conv   & $E_{MAC} \cdot FL_{Conv}$   &  $E_{MAC}\cdot  T  \cdot R \cdot  FL_{Conv} $ \\
           &  Other Conv  & $E_{MAC} \cdot FL_{Conv}$   & $E_{AC}  \cdot T  \cdot R \cdot FL_{Conv} $ \\
           \midrule
         \multirow{5}{*}{Self-Attention} &  $Q, K, V$ &  $E_{MAC} \cdot 3ND^2 $ &$E_{AC} \cdot T  \cdot R \cdot 3FL_{Conv} $  \\
         &  $f(Q, K, V)$ &  $E_{MAC} \cdot 2N^2D $   & $E_{AC} \cdot T  \cdot R \cdot2 ND^2 $  \\
         &  Scale & $E_{M} \cdot N^2$  &  - \\
         &  Softmax & $ E_{MAC} \cdot 2N^2 $ & -  \\
         &  Linear &  $E_{MAC} \cdot FL_{MLP}$ &  $E_{AC}  \cdot T  \cdot R \cdot FL_{MLP0} $ \\
        \midrule
            \multirow{2}{*}{MLP}   & Layer1   & $E_{MAC} \cdot FL_{MLP1}$   &  $E_{AC}\cdot  T  \cdot R \cdot  FL_{MLP1} $ \\
           &  Layer2  & $E_{MAC} \cdot FL_{MLP2}$   & $E_{AC}  \cdot T  \cdot R \cdot FL_{MLP2} $ \\

        \bottomrule
    \end{tabular}
\end{table*}

\clearpage
\onecolumn
\begin{longtable}{c|cccccc}

    \caption{Layer spiking firing rates of SpikeTrack-B256-T3 Search Branch.}
    \label{tab:spikingtate_search}\\
        \toprule
         & &  &  & T1  & T2 & T3 \\
         \cline{1-7}
     \multirow{2}{*}{Stage 1}   &\multicolumn{2}{c}{DownSampling} & Conv  &1  & -  & -  \\
     \cline{2-7}
      & \multirow{5}{*}{ConvFormer Spike Block} &   \multirow{3}{*}{SepSpikeConv} & PWConv1 & 1.4643  & - & - \\
      & &  &  DWConv&  1.3569 & - & - \\
      & &  & PWConv2 &  1.2556  & - & - \\
      & &\multirow{2}{*}{ Channel Conv}  & Conv1 & 1.5301  & - & - \\
      & &  & Conv2 & 0.2497  & - & - \\
      
        \cline{1-7}
           \multirow{2}{*}{Stage 2}   &\multicolumn{2}{c}{DownSampling} & Conv  &1.0722  & -  & -  \\

     \cline{2-7}
     
        & \multirow{11}{*}{Memory Retrieval Module} &    & Head-${Q}$ & 0.5670&-&	- \\
 & &  & ${Q_S}$1 &0.3496&0.3841&	0.3786\\ 
  & & \multirow{2}{*}{SepSpikeConv} &PWConv &  1.1416&	1.1947&	1.0329 \\
  & &    & DWConv & 0.3178&	0.3532&	0.3461\\
 & &  Project1  &Conv& 0.4639&	0.4746&	0.4770\\ 
  & &    &${Q_S}$2& 0.6606&	0.7157&	0.7218\\ 
  & & \multirow{2}{*}{TemporalFusion}   &WeightMLP1& 1.7143&	1.6588&	1.4170 \\
    & &    &WeightMLP2& 0.5168& 	0.5279& 	0.4708 \\
  & &  Project2  &Conv&  0.3493& 	-& 	- \\ 
      & &\multirow{2}{*}{ Channel MLP}  & Linear1 & 0.6840  & - & - \\
      & &  & Linear2 & 0.2197  & - & - \\

     \cline{2-7}
     
      & \multirow{5}{*}{ConvFormer Spike Block} &   \multirow{3}{*}{SepSpikeConv} & PWConv1 & 0.6638  & - & - \\
      & &  &  DWConv& 0.9398  & - & - \\
      & &  & PWConv2 & 0.6144  & - & - \\
      & &\multirow{2}{*}{ Channel Conv}  & Conv1 & 0.9048  & - & - \\
      & &  & Conv2 & 0.1488  & - & - \\
        \cline{1-7}

     \multirow{2}{*}{Stage 3}   &\multicolumn{2}{c}{DownSampling} & Conv  &0.7663  & -  & -  \\
     \cline{2-7}

        & \multirow{11}{*}{Memory Retrieval Module} &    & Head-${Q}$ & 0.6318&-&	-

 \\
 & &  & ${Q_S}$1 &0.2818&	0.3085&7	0.2999\\ 
  & & \multirow{2}{*}{SepSpikeConv} &PWConv &  0.8162&	0.9758&	0.8826 \\
  & &    & DWConv & 0.3054& 	0.3400&	0.3350 \\
 & &  Project1  &Conv& 0.3661& 	0.3896& 	0.4193\\ 
  & &    &${Q_S}$2& 0.3456&	0.3681&	0.3672\\ 
  & & \multirow{2}{*}{TemporalFusion}   &WeightMLP1& 0.7749&	0.8173&	0.6948\\
    & &    &WeightMLP2& 0.3484&	0.3827&	0.3497\\
  & &  Project2  &Conv&  0.1389& 	-& 	- \\ 
      & &\multirow{2}{*}{ Channel MLP}  & Linear1 & 0.6861  & - & - \\
      & &  & Linear2 & 0.1752  & - & - \\
      
     \cline{2-7}

      & \multirow{5}{*}{ConvFormer Spike Block} &   \multirow{3}{*}{SepSpikeConv} & PWConv1 & 0.6677  & - & - \\
      & &  &  DWConv& 0.9180  & - & - \\
      & &  & PWConv2 & 0.5269  & - & - \\
      & &\multirow{2}{*}{ Channel Conv}  & Conv1 &0.6199  & - & - \\
      & &  & Conv2 & 0.0906  & - & - \\
        \cline{2-7}     
            & \multirow{5}{*}{ConvFormer Spike Block} &   \multirow{3}{*}{SepSpikeConv} & PWConv1 & 0.9326  & - & - \\
      & &  &  DWConv& 0.5476  & - & - \\
      & &  & PWConv2 & 0.4813  & - & - \\
      & &\multirow{2}{*}{ Channel Conv}  & Conv1 & 0.6739  & - & - \\
      & &  & Conv2 & 0.065  & - & - \\
        \cline{1-7}

          \multirow{2}{*}{Stage 4}   &\multicolumn{2}{c}{DownSampling} & Conv  & 0.8766 & -  & -  \\

     \cline{2-7}

        & \multirow{5}{*}{Memory Retrieval Module} &    & Head-${Q}$ & 0.4919&-&	-

 \\
 & &  & ${Q_S}$1 &0.3510&	0.3837&	0.3756\\ 
  & & \multirow{2}{*}{SepSpikeConv} &PWConv &  2.1617&	2.3470&	2.3399 \\
  & &    & DWConv & 0.3806&	0.4072&0.4232\\
 & &  Project1  &Conv& 0.4583& 	0.4769& 	0.5183\\ 
  & &    &${Q_S}$2&0.7903&	0.8414&	0.8667\\ 
  & & \multirow{2}{*}{TemporalFusion}   &WeightMLP1& 2.8883& 	2.9805& 3.0267\\
    & &    &WeightMLP2& 0.2744&	0.2848&	0.2924\\
  & &  Project2  &Conv&  0.7324& 	-& 	- \\ 
      & &\multirow{2}{*}{ Channel MLP}  & Linear1 & 0.5211  & - & - \\
      & &  & Linear2 & 0.1942  & - & - \\
      
     \cline{2-7}

      & \multirow{10}{*}{TransFormer Spike Block} &   \multirow{3}{*}{SepSpikeConv} & PWConv1 &0.5216   & - & - \\
      & &  &  DWConv& 0.8427  & - & - \\
      & &  & PWConv2 & 0.3655  & - & - \\
      & & \multirow{5}{*}{SSA} & Head-$QKV$ & 0.5699  & - & - \\
& &  & $Q_S$ & 1.0672  & - & - \\
& &  & $K_S$ & 0.2967  & - & - \\
& &  & $V_S$ & 0.7571  & - & - \\
& &  & Linear & 2.1776  & - & - \\
      & &\multirow{2}{*}{ Channel MLP}  & Linear1 & 0.6194  & - & - \\
      & &  & Linear2 & 0.1132  & - & - \\
        \cline{2-7}   
  
            & \multirow{10}{*}{TransFormer Spike Block} &   \multirow{3}{*}{SepSpikeConv} & PWConv1 & 0.7469  & - & - \\
      & &  &  DWConv& 0.7103  & - & - \\
      & &  & PWConv2 & 0.4588  & - & - \\
      & & \multirow{5}{*}{SSA} & Head-$QKV$ & 0.6988  & - & - \\
& &  & $Q_S$ &0.7085  & - & - \\
& &  & $K_S$ & 0.2105  & - & - \\
& &  & $V_S$ & 0.3949  & - & - \\
& &  & Linear &1.3788  & - & - \\
      
      & &\multirow{2}{*}{ Channel MLP}  & Linear1 & 0.7699  & - & - \\
      & &  & Linear2 & 0.1025  & - & - \\
        \cline{2-7}

                    & \multirow{4}{*}{TransFormer Spike Block} &   \multirow{3}{*}{SepSpikeConv} & PWConv1 & 0.7743  & - & - \\
      & &  &  DWConv& 0.5849  & - & - \\
      & &  & PWConv2 & 0.3319  & - & - \\
            & &  & Head-$QKV$ & 0.7370  & - & - \\
& & \multirow{4}{*}{SSA} & $Q_S$ & 0.7367  & - & - \\
& &  & $K_S$ & 0.1592  & - & - \\
& &  & $V_S$ & 0.5306  & - & - \\
& &  & Linear & 1.2955  & - & - \\
      
      & &\multirow{2}{*}{ Channel MLP}  & Linear1 & 0.8817  & - & - \\
      & &  & Linear2 & 0.0799  & - & - \\
        \cline{2-7}

        & \multirow{11}{*}{Memory Retrieval Module} &    & Head-${Q}$ & 0.8867&-&	-

 \\
 & &  & ${Q_S}$1 &0.3233&	0.3501&	0.3357\\ 
  & & \multirow{2}{*}{SepSpikeConv} &PWConv & 1.2107&	1.4589&	1.424\\
  & &    & DWConv & 0.3771&	0.4003&	0.3844 \\
 & &  Project1  &Conv& 0.4302&	0.4495&	0.4991\\ 
  & &    &${Q_S}$2&0.5933&	0.6266&	0.6632\\ 
  & & \multirow{2}{*}{TemporalFusion}   &WeightMLP1& 1.7760&	1.9401&	2.0085
\\
    & &    &WeightMLP2& 1.0628&	1.1578&	1.1791\\
  & &  Project2  &Conv&  0.4434& 	-& 	- \\ 
      & &\multirow{2}{*}{ Channel MLP}  & Linear1 & 0.9183  & - & - \\
      & &  & Linear2 & 0.2099  & - & - \\
      
     \cline{2-7}

                    & \multirow{10}{*}{TransFormer Spike Block} &   \multirow{3}{*}{SepSpikeConv} & PWConv1 & 0.9292  & - & - \\
      & &  &  DWConv& 0.6615  & - & - \\
      & &  & PWConv2 & 0.3422  & - & - \\
       & &  & Head-$QKV$ & 0.8495  & - & - \\
& & \multirow{4}{*}{SSA} & $Q_S$ & 0.7161  & - & - \\
& &  & $K_S$ & 0.1350  & - & - \\
& &  & $V_S$ & 0.4256  & - & - \\
& &  & Linear & 1.0989  & - & - \\
      
      & &\multirow{2}{*}{ Channel MLP}  & Linear1 & 0.9172  & - & - \\
      & &  & Linear2 & 0.0677  & - & - \\
        \cline{2-7}

                    & \multirow{10}{*}{TransFormer Spike Block} &   \multirow{3}{*}{SepSpikeConv} & PWConv1 & 0.8158  & - & - \\
      & &  &  DWConv& 0.6410  & - & - \\
      & &  & PWConv2 & 0.345  & - & - \\
       & &  & Head-$QKV$ & 0.7505  & - & - \\
& & \multirow{4}{*}{SSA} & $Q_S$ & 0.7999  & - & - \\
& &  & $K_S$ & 0.1222  & - & - \\
& &  & $V_S$ & 0.3271  & - & - \\
& &  & Linear & 0.8896  & - & - \\
      
      & &\multirow{2}{*}{ Channel MLP}  & Linear1 & 0.9085  & - & - \\
      & &  & Linear2 &0.0744  & - & - \\
        \cline{2-7}

                    & \multirow{10}{*}{TransFormer Spike Block} &   \multirow{3}{*}{SepSpikeConv} & PWConv1 & 0.7901  & - & - \\
      & &  &  DWConv& 0.6209  & - & - \\
      & &  & PWConv2 & 0.2965  & - & - \\
       & &  & Head-$QKV$ & 0.7643  & - & - \\
& & \multirow{4}{*}{SSA} & $Q_S$ & 0.7730  & - & - \\
& &  & $K_S$ & 0.1993  & - & - \\
& &  & $V_S$ & 0.5176 & - & - \\
& &  & Linear & 1.6896  & - & - \\
      
      & &\multirow{2}{*}{ Channel MLP}  & Linear1 & 0.8989  & - & - \\
      & &  & Linear2 & 0.0871  & - & - \\
        \cline{1-7}

     \multirow{2}{*}{Stage 5}   &\multicolumn{2}{c}{DownSampling} & Conv  & 0.5798 & -  & -  \\
     \cline{2-7}

        & \multirow{11}{*}{Memory Retrieval Module} &    & Head-${Q}$ & 1.0431&-&-\\
 & &  & ${Q_S}$1 &0.4587&	0.4759&	0.4673 \\ 
  & & \multirow{2}{*}{SepSpikeConv} &PWConv & 1.5555&	1.5786&	1.4752\\
  & &    & DWConv & 0.3697&	0.3760&	0.3707  \\
 & &  Project1  &Conv& 0.5233&	0.5438&	0.5847\\ 
  & &    &${Q_S}$2&0.3806&	0.4307&	0.4813 \\ 
  & & \multirow{2}{*}{TemporalFusion}   &WeightMLP1& 1.2338&	1.489&	1.5396\\
    & &    &WeightMLP2& 0.9094&	1.0829&	1.1119\\
  & &  Project2  &Conv& 0.2615& 	-& 	- \\ 
      & &\multirow{2}{*}{ Channel MLP}  & Linear1 & 1.0764  & - & - \\
      & &  & Linear2 & 0.1975 & - & - \\
      
     \cline{2-7}

      & \multirow{10}{*}{TransFormer Spike Block} &   \multirow{3}{*}{SepSpikeConv} & PWConv1 &1.1655   & - & - \\
      & &  &  DWConv& 0.5649  & - & - \\
      & &  & PWConv2 & 0.5767  & - & - \\
      & & \multirow{5}{*}{SSA} & Head-$QKV$ & 0.9361  & - & - \\
& &  & $Q_S$ & 0.7536  & - & - \\
& &  & $K_S$ & 0.0724  & - & - \\
& &  & $V_S$ & 0.1860  & - & - \\
& &  & Linear & 0.5488  & - & - \\
      & &\multirow{2}{*}{ Channel MLP}  & Linear1 & 1.0615  & - & - \\
      & &  & Linear2 & 0.0453  & - & - \\
        \cline{2-7}   
                    & \multirow{10}{*}{TransFormer Spike Block} &   \multirow{3}{*}{SepSpikeConv} & PWConv1 & 0.8718  & - & - \\
      & &  &  DWConv& 0.2728  & - & - \\
      & &  & PWConv2 & 0.2346  & - & - \\
       & &  & Head-$QKV$ & 0.4639  & - & - \\
& & \multirow{4}{*}{SSA} & $Q_S$ & 1.0436  & - & - \\
& &  & $K_S$ & 0.0690  & - & - \\
& &  & $V_S$ & 0.0402  & - & - \\
& &  & Linear & 0.2674  & - & - \\
      
      & &\multirow{2}{*}{ Channel MLP}  & Linear1 & 0.1904  & - & - \\
      & &  & Linear2 &0.0043  & - & - \\

     \cline{2-7}

        & \multirow{11}{*}{Memory Retrieval Module} &    & Head-${Q}$ & 0.2893&-&-\\
 & &  & ${Q_S}$1 &0.2288&	0.2511&	0.2457\\ 
  & & \multirow{2}{*}{SepSpikeConv} &PWConv & 1.4003&	1.5141&	1.4140\\
  & &    & DWConv & 0.2876& 	0.3126&	0.3128\\
 & &  Project1  &Conv& 0.3465& 	0.3634& 	0.3737\\ 
  & &    &${Q_S}$2&0.5225&	0.5534&	0.5321\\ 
  & & \multirow{2}{*}{TemporalFusion}   &WeightMLP1& 2.3932&	2.5136&	2.3640\\
    & &    &WeightMLP2& 0.7827&	0.8143&	0.7502\\
  & &  Project2  &Conv& 0.5717& 	-& 	- \\ 
      & &\multirow{2}{*}{ Channel MLP}  & Linear1 & 0.3105  & - & - \\
      & &  & Linear2 & 0.2024 & - & - \\

        \cline{1-7}

 \multirow{15}{*}{Head}& \multirow{5}{*}{Location}  &    & Conv1 & 0.3149  & - & - \\
  &                      &                             & Conv2 & 0.2198  & - & - \\
  &                      &                             & Conv3 & 0.3037  & - & - \\
  &                      &                             & Conv4 & 0.6767  & - & - \\
  &                      &                             & Conv5 & 1.8021  & - & - \\
\cline{2-7}
& \multirow{5}{*}{Offset} &    & Conv1 & 0.3149  & - & - \\
  &                      &                             & Conv2 & 0.2182  & - & - \\
  &                      &                             & Conv3 & 0.3431  & - & - \\
  &                      &                             & Conv4 & 0.5206  & - & - \\
  &                      &                             & Conv5 & 0.6281 & - & - \\
\cline{2-7}
& \multirow{5}{*}{Size}  &                            &  Conv1 & 0.3149  & - & - \\
  &                      &                             & Conv2 & 0.2271  & - & - \\
  &                      &                             & Conv3 & 0.3000  & - & - \\
  &                      &                             & Conv4 & 0.6038  & - & - \\
  &                      &                             & Conv5 & 0.5048  & - & - \\  
\bottomrule

\end{longtable}

\begin{longtable}{c|cccccc}

    \caption{Layer spiking firing rates of SpikeTrack-B256-T3 Template Branch.}
    \label{tab:spikingtate_template}\\
        \toprule
         & &  &  & T1  & T2 & T3 \\
         \cline{1-7}
     \multirow{2}{*}{Stage 1}   &\multicolumn{2}{c}{DownSampling} & Conv  &1  & 1  & 1  \\
     \cline{2-7}
      & \multirow{5}{*}{ConvFormer Spike Block} &   \multirow{3}{*}{SepSpikeConv} & PWConv1 & 1.458  & 1.4853 & 1.5083 \\
      & &  &  DWConv&  1.3534 & 1.3790 & 1.3954 \\
      & &  & PWConv2 &  1.2563  & 1.3188 & 1.3337 \\
      & &\multirow{2}{*}{ Channel Conv}  & Conv1 & 1.5297  & 1.5035 & 1.5087 \\
      & &  & Conv2 & 0.2500  & 0.2496 & 0.2460 \\
      
              \cline{1-7}
           \multirow{2}{*}{Stage 2}   &\multicolumn{2}{c}{DownSampling} & Conv  &1.0752  & 1.0885  & 1.0787  \\
\cline{2-7}

    & \multirow{3}{*}{Memory Retrieval Module} &    & Head-${KV}$ & 0.5344  & 0.5987 & 0.5938 \\
      & &  &  $V_S$&  0.3571 & 0.3961 & 0.3985 \\
      & &  & $K_S$ &  0.2998  & 0.3259 & 0.3188 \\

\cline{2-7}
     
      & \multirow{5}{*}{ConvFormer Spike Block} &   \multirow{3}{*}{SepSpikeConv} & PWConv1 & 0.6596  & 0.6715 & 0.6727 \\
      & &  &  DWConv& 0.9408  & 0.9589 & 0.9664 \\
      & &  & PWConv2 & 0.6114  & 0.6465 & 0.6431 \\
      & &\multirow{2}{*}{ Channel Conv}  & Conv1 & 0.9114  & 0.9267 & 0.9209 \\
      & &  & Conv2 & 0.1480  & 0.1460 & 0.1465 \\
      \cline{2-7}

\cline{1-7}

     \multirow{2}{*}{Stage 3}   &\multicolumn{2}{c}{DownSampling} & Conv  &0.7826  & 0.7730  & 0.7694  \\
     \cline{2-7}

    & \multirow{3}{*}{Memory Retrieval Module} &    & Head-${KV}$ & 0.6126  & 0.6834 & 0.6718 \\
      & &  &  $V_S$&  0.3057 &0.3386 & 0.3369 \\
      & &  & $K_S$ &  0.3060  & 0.3400 & 0.3313 \\

    \cline{2-7}
     
      & \multirow{5}{*}{ConvFormer Spike Block} &   \multirow{3}{*}{SepSpikeConv} & PWConv1 & 0.6688  & 0.6766 & 0.6739 \\
      & &  &  DWConv& 0.9315  & 0.9402 & 0.9380 \\
      & &  & PWConv2 & 0.5135  & 0.5246 & 0.5213 \\
      & &\multirow{2}{*}{ Channel Conv}  & Conv1 &0.6318  & 0.6374 & 0.6359 \\
      & &  & Conv2 & 0.0906  & 0.0903 & 0.0888 \\
        \cline{2-7}     
            & \multirow{5}{*}{ConvFormer Spike Block} &   \multirow{3}{*}{SepSpikeConv} & PWConv1 & 0.9604  & 0.9774 & 0.9793 \\
      & &  &  DWConv& 0.5758  & 0.5904 & 0.5886 \\
      & &  & PWConv2 &0.4773  & 0.4883 & 0.4884 \\
      & &\multirow{2}{*}{ Channel Conv}  & Conv1 & 0.6978  & 0.6975 & 0.7002 \\
      & &  & Conv2 & 0.0694  & 0.0680 & 0.0678 \\
        \cline{1-7}

          \multirow{2}{*}{Stage 4}   &\multicolumn{2}{c}{DownSampling} & Conv  & 0.9081 & 0.9001 & 0.8983  \\
     \cline{2-7}
     
        & \multirow{3}{*}{Memory Retrieval Module} &    & Head-${KV}$ & 0.4755  & 0.5012 & 0.5137 \\
      & &  &  $V_S$&  0.4067 &0.4387 & 0.4539 \\
      & &  & $K_S$ &  0.5307  & 0.5693 & 0.5842 \\

    \cline{2-7}
     
      & \multirow{10}{*}{TransFormer Spike Block} &   \multirow{3}{*}{SepSpikeConv} & PWConv1 &0.4915   & 0.5026 & 0.5066 \\
      & &  &  DWConv& 0.8554  & 0.8717 & 0.8803 \\
      & &  & PWConv2 &0.3502  & 0.3697 & 0.3749 \\
      & & \multirow{5}{*}{SSA} & Head-$QKV$ & 0.5477  & 0.5811 & 0.5833 \\
& &  & $Q_S$ & 1.1002  & 1.2265 & 1.2395 \\
& &  & $K_S$ & 0.2663  & 0.2994 & 0.3055 \\
& &  & $V_S$ & 0.7438  & 0.7834 & 0.7932 \\
& &  & Linear & 2.1391  & 2.3258 & 2.3621 \\
      & &\multirow{2}{*}{ Channel MLP}  & Linear1 & 0.6040  & 0.5976 & 0.5933 \\
      & &  & Linear2 & 0.0985  & 0.0980 & 0.0974 \\
        \cline{2-7}   
  
            & \multirow{10}{*}{TransFormer Spike Block} &   \multirow{3}{*}{SepSpikeConv} & PWConv1 & 0.7451  & 0.7327 & 0.7315 \\
      & &  &  DWConv& 0.7205  & 0.7095 & 0.7085 \\
      & &  & PWConv2 & 0.4378  & 0.4552 & 0.4575 \\
      & & \multirow{5}{*}{SSA} & Head-$QKV$ & 0.7029  & 0.7192 & 0.7023 \\
& &  & $Q_S$ &0.7605  & 0.7370 & 0.7062 \\
& &  & $K_S$ & 0.2392  & 0.2287 & 0.2077 \\
& &  & $V_S$ & 0.4056  & 0.4129 & 0.4003 \\
& &  & Linear &1.6298  & 1.5722 & 1.3865 \\
      
      & &\multirow{2}{*}{ Channel MLP}  & Linear1 & 0.7850&	0.7871	&0.7767
 \\
      & &  & Linear2 & 0.1052&	0.1089&	0.1039
 \\
        \cline{2-7}

                    & \multirow{4}{*}{TransFormer Spike Block} &   \multirow{3}{*}{SepSpikeConv} & PWConv1 & 0.7683&	0.7498&	0.7574 \\
      & &  &  DWConv& 0.5794&	0.5768&	0.5805 \\
      & &  & PWConv2 & 0.3307&	0.3437&	0.3452  \\
            & &  & Head-$QKV$ & 0.7291&	0.7302&	0.7277 \\
& & \multirow{4}{*}{SSA} & $Q_S$ & 0.7556&	0.7416&	0.7483 \\
& &  & $K_S$ & 0.1557&	0.1395&	0.1411 \\
& &  & $V_S$ & 0.5425&	0.5424& 	0.5408 \\
& &  & Linear & 1.3957&	1.3158&	1.3046 \\
      
      & &\multirow{2}{*}{ Channel MLP}  & Linear1 & 0.8744&	0.8590&	0.8692 \\
      & &  & Linear2 & 0.0735&	0.0722&	0.0723 \\

     \cline{2-7}
     
        & \multirow{3}{*}{Memory Retrieval Module} &    & Head-${KV}$ & 0.8776&	0.8964&	0.9105
 \\
      & &  &  $V_S$&  0.2765&	0.3190&	0.3222  \\
      & &  & $K_S$ &  0.2897&	0.2831&	0.2953 \\

    \cline{2-7}

                    & \multirow{10}{*}{TransFormer Spike Block} &   \multirow{3}{*}{SepSpikeConv} & PWConv1 &0.8760&	0.8442&	0.8623 \\
      & &  &  DWConv& 0.6305&	0.6225&	0.6328 \\
      & &  & PWConv2 & 0.3151&	0.3138&	0.3219 \\
       & &  & Head-$QKV$ & 0.8052&	0.8112&	0.8115 \\
& & \multirow{4}{*}{SSA} & $Q_S$ & 0.6392&	0.6850&	0.6665\\
& &  & $K_S$ & 0.1438&	0.1365&	0.1354 \\
& &  & $V_S$ & 0.4303&	0.4325&	0.4365  \\
& &  & Linear & 1.1670&	1.2427& 	1.2174\\
      
      & &\multirow{2}{*}{ Channel MLP}  & Linear1 & 0.9013&	0.8820&	0.8989\\
      & &  & Linear2 & 0.0616&	0.0599&	0.0602\\
        \cline{2-7}

        & \multirow{10}{*}{TransFormer Spike Block} &   \multirow{3}{*}{SepSpikeConv} & PWConv1 & 0.7866&	0.7636&	0.7793\\
      & &  &  DWConv& 0.6259&	0.6263&	0.6347\\
      & &  & PWConv2 & 0.3305&0.3450	&0.3529 \\
       & &  & Head-$QKV$ & 0.7390&	0.7483&	0.7388\\
       
& & \multirow{4}{*}{SSA} & $Q_S$ & 0.7814&	0.8434&	0.8010 \\
& &  & $K_S$ & 0.1488&	0.1365&	0.1227 \\
& &  & $V_S$ & 0.3419&	0.3411&	0.3399 \\
& &  & Linear &0.9959&	1.0392&	0.9468 \\
      
      & &\multirow{2}{*}{ Channel MLP}  & Linear1 & 0.8848&	0.8524&	0.8672\\
      & &  & Linear2 &0.0808&	0.0799&	0.0751\\
        \cline{2-7}

    & \multirow{10}{*}{TransFormer Spike Block} &   \multirow{3}{*}{SepSpikeConv} & PWConv1 & 0.7143&	0.6833&	0.7108 \\
      & &  &  DWConv& 0.5917&	0.5949&	0.5955\\
      & &  & PWConv2 & 0.2799&	0.2859&	0.2897 \\
       & &  & Head-$QKV$ & 0.7315&	0.7115&	0.7250 \\
& & \multirow{4}{*}{SSA} & $Q_S$ & 0.8245&	0.8770&	0.8687\\
& &  & $K_S$ & 0.1935&	0.1864&	0.1898\\
& &  & $V_S$ & 0.4987&	0.4917&	0.5064\\
& &  & Linear & 1.8150& 	1.8574&	1.8907\\
      
      & &\multirow{2}{*}{ Channel MLP}  & Linear1 & 0.8072&	0.7669&	0.8002\\
      & &  & Linear2 & 0.1164&	0.1183&	0.1128\\
        \cline{1-7}

     \multirow{2}{*}{Stage 5}   &\multicolumn{2}{c}{DownSampling} & Conv  & 0.3914&0.3706&	0.3886\\

     \cline{2-7}
     
        & \multirow{3}{*}{Memory Retrieval Module} &    & Head-${KV}$ & 1.0682&1.0843&	1.0665
\\
      & &  &  $V_S$&  0.1857&	0.1970&	0.1987\\
      & &  & $K_S$ &  0.2269&	0.2089&	0.2198 \\

     \cline{2-7}
      & \multirow{10}{*}{TransFormer Spike Block} &   \multirow{3}{*}{SepSpikeConv} & PWConv1 &1.0331&	1.0798&	1.0915 \\
      & &  &  DWConv& 0.4430&	0.4637	&0.4686 \\
      & &  & PWConv2 & 0.3645&	0.4348&	0.4429\\
      & & \multirow{5}{*}{SSA} & Head-$QKV$ & 0.8320&	0.8420&	0.8145\\
& &  & $Q_S$ & 0.9074&	0.9703&	1.0095 \\
& &  & $K_S$ & 0.0435&	0.0389&	0.0375\\
& &  & $V_S$ & 0.1377&	0.1324&	0.1297 \\
& &  & Linear & 0.5249&	0.4881&	0.4730 \\
      & &\multirow{2}{*}{ Channel MLP}  & Linear1 & 0.9349&0.9175&0.9127\\
      & &  & Linear2 & 0.0181&	0.01697&	0.0161\\
        \cline{2-7}   
                    & \multirow{10}{*}{TransFormer Spike Block} &   \multirow{3}{*}{SepSpikeConv} & PWConv1 & 0.7944&	0.7935&	0.7932\\
      & &  &  DWConv& 0.1714& 	0.1775& 	0.1778\\
      & &  & PWConv2 & 0.1576&	0.1701&	0.1740 \\
       & &  & Head-$QKV$ & 0.3908&	0.3597&	0.3534\\
& & \multirow{4}{*}{SSA} & $Q_S$ & 0.5916&	0.6136&	0.6324 \\
& &  & $K_S$ &0.0311&	0.0302&	0.0299 \\
& &  & $V_S$ & 0.0268	&0.0255&0.0246\\
& &  & Linear & 0.0421&	0.0446&	0.0520\\
      
      & &\multirow{2}{*}{ Channel MLP}  & Linear1 & 0.3862&	0.3311&	0.3099\\
      & &  & Linear2 &0.0091&	0.0077&	0.0068\\

     \cline{2-7}
     
        & \multirow{3}{*}{Memory Retrieval Module} &    & Head-${KV}$ &0.8044&0.7835&	0.7841

\\
      & &  &  $V_S$& 0.3875&	0.3687&	0.3515\\
      & &  & $K_S$ &  0.3411&	0.3344&	0.3173
 \\

    \bottomrule

\end{longtable}

\twocolumn

\end{document}